\documentclass[twocolumn,10pt]{article}

\usepackage[utf8]{inputenc}
\usepackage{amsmath,amsfonts}
\usepackage{amsthm}
\usepackage{amssymb}
\usepackage{mathtools}
\usepackage{bbm}

\usepackage{graphicx}
\usepackage{booktabs}
\usepackage{hyperref}
\usepackage{mathtools}
\usepackage[subtle]{savetrees}
\usepackage{color}

\usepackage{geometry}
\usepackage{helvet}
\usepackage[T1]{fontenc}

\usepackage[superscript]{cite}

\usepackage{abstract}
\newtheorem{theorem}{Theorem}
\newtheorem{lemma}{Lemma}

\newtheorem{definition}{Definition}

\newcommand{\squishlist}{
   \begin{list}{$\bullet$}
    { \setlength{\itemsep}{2pt}    \setlength{\parsep}{0pt}
      \setlength{\topsep}{5pt}     \setlength{\partopsep}{0pt}
      \setlength{\leftmargin}{1.35em} \setlength{\labelwidth}{1em}
      \setlength{\labelsep}{0.5em} } }

\newcommand{\squishend}{
    \end{list}  }

\newcommand{\acronym}{\textsc{DeepDECS}}

\usepackage{caption}
\usepackage{subcaption}

\usepackage{tikz}
\usepackage{adjustbox}
\usepackage{pgfplots}
\pgfplotsset{
  compat=1.12
}
\usetikzlibrary{patterns}

\begin{document}

\title{Discrete-Event Controller Synthesis for Autonomous Systems\\ with Deep-Learning Perception Components}
\author{Radu Calinescu\thanks{Department of Computer Science, University of York, York, UK.}, Calum Imrie\footnotemark[1], Ravi Mangal\thanks{Carnegie Mellon University, Silicon Valley, USA}, Gena\'{i}na Nunes Rodrigues\thanks{Department of Computer Science, University of Bras\'{i}lia, Brazil},\\ Corina P\u{a}s\u{a}reanu\footnotemark[2], Misael Alpizar Santana\footnotemark[1], and Gricel V\'{a}zquez\footnotemark[1]}
\date{}

\twocolumn[
  \begin{@twocolumnfalse}
    \maketitle
    \begin{abstract}
    
    \vspace*{-3mm}
    \textbf{
    We present \acronym, a new method for the synthesis of correct-by-construction discrete-event controllers for autonomous systems that use deep neural network (DNN) classifiers for the perception step of their decision-making processes. Despite major advances in deep learning in recent years, providing safety guarantees for these systems remains very challenging. Our controller synthesis method addresses this challenge by integrating DNN verification with the synthesis of verified Markov models. The synthesised models correspond to discrete-event controllers guaranteed to satisfy the safety, dependability and performance requirements of the autonomous system, and to be Pareto optimal with respect to a set of optimisation objectives. We use the method in simulation to synthesise controllers for mobile-robot collision mitigation and for maintaining driver attentiveness in shared-control autonomous driving.
    }
    \end{abstract}
    
    \vspace*{1mm}
  \end{@twocolumnfalse}
  ]
  \saythanks
  
\maketitle

Autonomous systems (AS) must perceive and adapt to changes in their environment. In application domains as diverse as medicine~\cite{chen-etal2020,de2018clinically}, finance~\cite{fischer2018deep} and transportation~\cite{grigorescu2020survey,tabernik2019deep}, this perception is often performed using deep neural network (DNN) classifiers. This integration of DNN perception into the AS control loop poses major assurance challenges.~\cite{10.1145/3453444} In particular, the long-established methods for formal software verification~\cite{d2008survey} cannot be used to provide safety and performance guarantees for AS comprising both traditional software and deep-learning components. Furthermore, verification methods developed specifically for DNNs focus on verifying robustness to changes in DNN inputs~\cite{katz2019marabou,10.1007/978-3-319-63387-9_1} or input clusters~\cite{gopinath2018deepsafe}, and are equally unable to provide system-level guarantees for the controllers of DNN-perception AS.

Our paper presents a controller synthesis method that addresses this significant limitation.  %To that end, 
Called \acronym\ (\underline{Deep} neural network perception \emph{aware} \underline{D}iscrete-\underline{E}vent \underline{C}ontroller \underline{S}ynthesis), our method employs a suite of DNN verification techniques to quantify the \emph{aleatory uncertainty} introduced by the use of DNN perception in the AS control loop. 
Discrete-event controllers guaranteed to satisfy the AS requirements are then synthesised from a stochastic model that captures this uncertainty \emph{and} leverages the high accuracy that DNNs can achieve for verified inputs. 

We start by introducing \acronym\ and its unique use of both DNN and traditional verification. Next, we describe the use of \acronym\ to devise controllers for mobile-robot collision mitigation and driver attentiveness management in vehicles equipped with Level 3 (conditional automation)
automated driving systems.\cite{SAE-J3016-2018} Finally, we discuss \acronym\ and its contributions in the context of related work.

%%%%%%%%%%%%%%%%%%%%%%%%%%%%%%%%%%%%%%%%%%%%%%
\subsection*{1 \acronym\ controller synthesis}
\label{sec:intro}

\noindent
\acronym\ models the design space of the AS controller under development as a \emph{parametric discrete-time Markov chain} (pDTMC). The uncertainty introduced by the
deep-learning perception and that inherent to the environment are modelled by the probabilities of transition between the states of this pDTMC. The controller synthesis problem involves finding combinations of parameter values for which the pDTMC satisfies strict safety, dependability and performance constraints, and is Pareto-optimal with respect to a set of optimisation objectives. These constraints and optimisation objectives are formalised as probabilistic temporal logic formulae. 

\begin{figure*}
    \centering 
    \begin{subfigure}[b]{\textwidth}
    \centering
    \includegraphics[width=0.85\hsize]{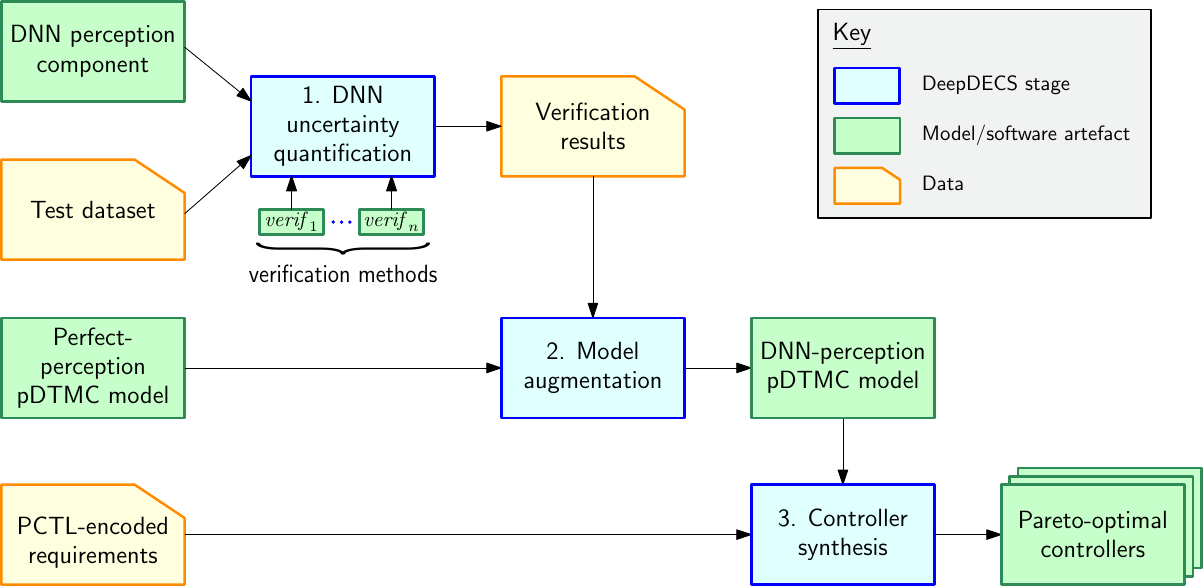}
    \caption{\acronym\ generates discrete-event controllers aware of the uncertainty induced by the DNN perception component of an AS in three stages. First, in a \emph{DNN uncertainty quantification} stage, $n$ DNN verification techniques are used to evaluate the DNN perception component over a test dataset representative for the operational design domain (ODD) of the AS. The verification results provide a quantification of the DNN prediction uncertainty within the system ODD. Next, the \emph{Model augmentation} stage uses these results---and an ideal-system pDTMC model that assumes perfect perception---to assemble a pDTMC system model that takes the DNN-induced uncertainty into account. Finally, the \emph{Controller synthesis} stage uses this pDTMC model to synthesise a set of Pareto-optimal discrete-event controllers guaranteed to satisfy the AS requirements (i.e., constraints and optimisation objectives) encoded in probabilistic computation tree logic (PCTL).
    }
    \end{subfigure}
    
    \vspace{4mm}
    \begin{subfigure}[b]{\textwidth}
    \centering
    \includegraphics[width=0.47\hsize]{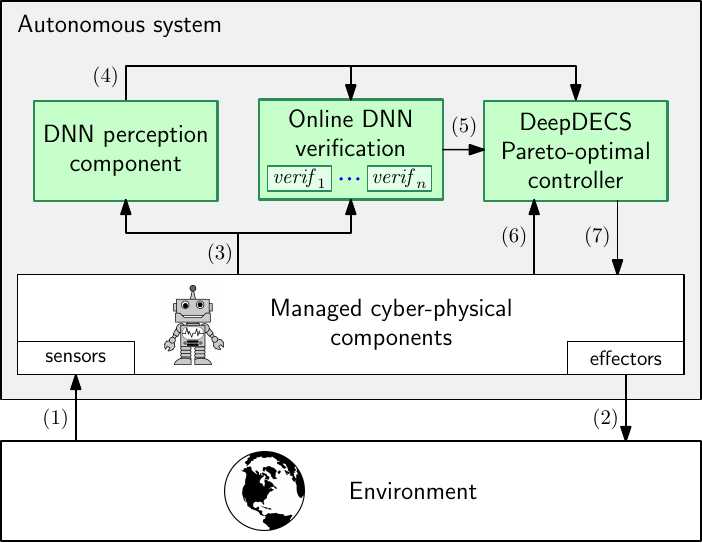}
    \caption{The cyber-physical components of an autonomous system managed by a \acronym\ controller monitor their environment through sensors~(1) and perform actions that affect it through effectors~(2). A DNN perception component uses a combination~(3) of preprocessed sensor data and data about these managed components to classify the state of the environment~(4). The $n$ verification techniques used for the \acronym\ controller synthesis are also applied to the classification~(4) and the DNN input~(3) that produced it. Using the online DNN verification results~(5) alongside the classification~(4) and additional state information~(6) obtained directly from the managed cyber-physical components, the \acronym\ controller updates~(7) the controllable parameters of these components in line with the system requirements. 
    }
    \end{subfigure}
    \caption{\acronym\ controller synthesis (a), and deployment (b)}
    \label{fig:approach}
\end{figure*}

\acronym\ derives the pDTMC underpinning its controller synthesis automatically from (i)~DNN verification results that quantify the uncertainty introduced by the DNN perception, and (ii)~an ``ideal'' pDTMC that models the AS behaviour assuming perfect perception (Figure~\ref{fig:approach}a). Pareto-optimal \acronym\ controllers are then synthesised by applying a combination of probabilistic model checking and search techniques to this pDTMC. As shown in Figure~\ref{fig:approach}b, the resulting controller operates by reacting to changes in the system, in the DNN outputs and, unique to \acronym, in the results of the online verification of the DNN classification. We detail the \acronym\ stages below.

%%%%%%%%%%%%%%%%%%%%%%%%%%%%%%%%%%
\medskip
\noindent
\textbf{Stage 1: DNN uncertainty quantification} 

\medskip
%\noindent
\emph{a) Verification of DNN classifiers.}
A $K$-class DNN classifier $f$ is a function that maps a $d$-dimensional input to a \emph{class} from the set $[K]=\{1,2,\ldots,K\}$:
\begin{equation}
  f:\mathbb{R}^d\rightarrow [K].
  \label{eq:classifier}
\end{equation}
DNN classifiers are learnt from data, and are not guaranteed to always classify their input correctly. DNN verification techniques (examples of which are provided in the Methods section) can help assess the quality of a classifier for a given input. A verification technique has the general form
\begin{equation}
  \label{eq:verif}
  \mathit{verif}:(\mathbb{R}^d\rightarrow [K]) \times \mathbb{R}^d\rightarrow \mathbb{B},
\end{equation}
such that, for a classifier $f \in \mathbb{R}^d\rightarrow [K]$ and an input $x\in \mathbb{R}^d$, $\mathit{verif}(f,x)=\mathsf{true}$ if the technique \emph{deems} the DNN $f$ likely to classify the input $x$ correctly, and $\mathit{verif}(f,x)=\mathsf{false}$ otherwise. 

\medskip
\emph{b) Uncertainty quantification.} DNN perception introduces aleatory uncertainty since DNNs cannot classify all inputs accurately. To quantify this uncertainty, \acronym~uses $n\geq 0$ DNN verification techniques $\mathit{verif}_1$, $\mathit{verif}_2$, \ldots, $\mathit{verif}_n$,  
and a test dataset $X\subset \mathbb{R}^d$ that represents a
\emph{statistically representative sample} of the inputs that the AS will encounter in operation. The $n$ verification techniques are used to partition $X$ into $2^n$ subsets comprising inputs $x$ with the same verification results. We note that using only a few verification techniques (e.g., $n\leq 3$) yields a small number of such subsets. Formally, given a DNN $f$, \acronym~constructs the subset
\begin{equation}
  \label{eq:subset}
  X_v
  = \{ x\in X \mid \mathit{verif}(f,x)=v \}
\end{equation}
for every $v=(v_1,v_2,\ldots,v_n)\in\mathbb{B}^n$, 
where $\mathit{verif}(f,x)=(\mathit{verif}_1(f,x),$ $\mathit{verif}_2(f,x),\ldots,\mathit{verif}_n(f,x))$.
We use each subset~\eqref{eq:subset} to obtain a $K\times K$ confusion matrix $\mathcal{C}_{v}$ such that, for any $k,k'\in [K]$, the element in row $k$ and column $k'$ of $\mathcal{C}_{v}$ represents the number of class-$k$ inputs from $X_{v}$ that the DNN classifies as belonging to class $k'$:
\begin{equation}
  \label{eq:confusion}
  \mathcal{C}_v
  [k,k'] = \#\left\{x\in X_v \mid  f^*(x)=k \wedge f(x)=k'\right\},
\end{equation}
where $f^*(x)$ represents the true class of $x$ and, for any set $A$, $\#A$ denotes its cardinality.

As $X$ is representative of the DNN inputs that the AS encounters in operation, we henceforth assume that the probability that a class-$k$ input $x$ is classified 
%by the DNN 
as belonging to class $k'$ when it satisfies $\mathit{verif}(f,x)=v$ is given by:
\begin{multline}
    \label{eq:DNN-probabilities}
    p_{kk'v}=\mathit{Pr}\left( f(x)=k' \:\wedge\: \mathit{verif}(f,x)=v \;\right|\left. f^*(x)=k \right)\\ = \frac{C_v[k,k']}
    {\sum_{v'\in\mathbb{B}^n} \sum_{k''\in[K]} C_{v'}[k,k'']}
    = \frac{C_v[k,k']}
    {\#\{x\in X\mid f^*(x)=k\}}.
\end{multline}
Formally, this results holds as $\#X\rightarrow\infty$. We note that 
\begin{equation}
  \sum_{(k',v)'\in[K]\times\mathbb{B}^n} p_{kk'v}=1.
\end{equation}

%%%%%%%%%%%%%%%%%%%%%%%%%%%%%%%%%%
\bigskip
\noindent
\textbf{Stage 2: Model augmentation} 

\medskip
%\noindent
\emph{a) Markov chains.} \acronym\ models the design space (i.e., the possible variants) for an AS controller as a pDTMC augmented with \emph{rewards}.  A reward-augmented discrete-time Markov chain (DTMC) over a set of atomic propositions $\mathit{AP}$ is a tuple 
\begin{equation}
  \mathcal{M}=(S, s_0, P, L, R),
  \label{eq:DTMC}
\end{equation}
where $S\neq\emptyset$ is a finite set of states; $s_0 \in S$ is the initial state; $P : S \times S \rightarrow [0,1]$ is a transition probability function such that, for any states $s,s'\in S$, $P(s,s')$ gives the probability of transition from state $s$ to state $s'$ and $\sum_{s'\in S}P(s,s')=1$;  $L: S \rightarrow 2^\mathit{AP}$ is a labelling function that maps every state $s \in S$ to the atomic propositions from $\mathit{AP}$ that hold in that state; and $R$ is a set of \emph{reward structures}, i.e., function pairs $(\rho,\iota)$ that associate non-negative values with the DTMC states and transitions:
\begin{equation} 
  \label{eq:reward-structure}
   \rho:S\rightarrow \mathbb{R}_{\geq0},\;\; \iota:S\times S\rightarrow  \mathbb{R}_{\geq0}.
\end{equation}
A pDTMC is a DTMC~\eqref{eq:DTMC} comprising transition probabilities and/or rewards specified as rational functions over a set of continuous variables,~\cite{Daws:2004:SPM:2102873.2102899} i.e., functions that can be written as fractions whose numerators and denominators are polynomial expressions,  e.g., $\frac{1-p_1}{p_2}$.

\acronym\ uses \emph{probabilistic computation tree logic} (PCTL)~\cite{Hansson1994,bianco1995model} extended with rewards~\cite{andova2003discrete} to formalise AS requirements. Reward-augmented PCTL (cf.~Methods) supports the specification of constraints such as `the robot should not incur a collision until done with its mission with probability at least 0.75' ($\mathcal{P}[\neg\mathsf{collision} \;\mathrm{U}\; \mathsf{done}]\geq 0.75$) and optimisation objectives such as `minimise the expected mission time' ($\textrm{minimise }\mathcal{R}^\mathsf{time} [\mathrm{F}\; done]$). 

\begin{figure*}[t]
     \centering
     \begin{subfigure}[c]{0.433\linewidth}
         \centering
         \includegraphics[width=1.0\linewidth]{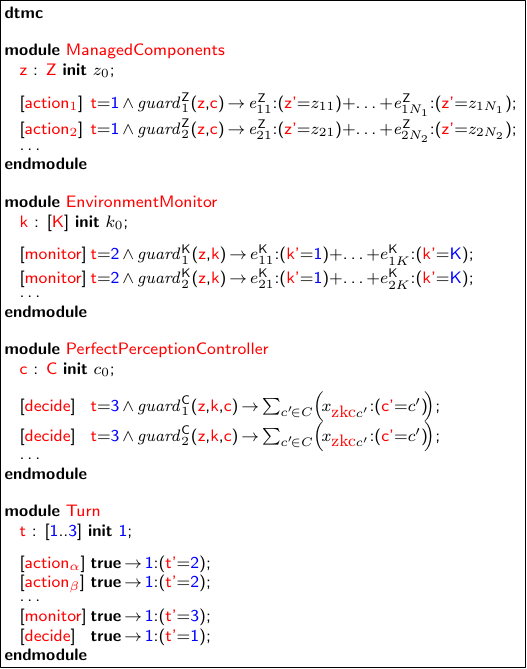}
         \caption{A \acronym\ perfect-perception pDTMC comprises four modules. The module \textcolor{red}{ManagedComponents} models the controlled components of the system by specifying how their state $z$ changes as a result of set of actions $\mathit{Act}=\{$\textcolor{red}{$\mathsf{action}_1$},\textcolor{red}{$\mathsf{action}_2$}$,\ldots\}$ performed when the value of the turn flag is $t=1$; the guards of this module depend only on the state $z$ of the components and the control parameters $c$. The module \textcolor{red}{EnvironmentMonitor} models how the evolving state of environment $k$ changes when observed through monitoring when the value of the turn flag is $t=2$; the guards of this module depend only on the state $z$ of the system components and on the current state $k$ of the environment. The controller decisions are defined by the \textcolor{red}{PerfectPerceptionController} module with parameters~\eqref{eq:ideal-controller-parameters}. Finally, the module \textcolor{red}{Turn} sets the turn flag to $t=2$ when, after specific actions \textcolor{red}{$\mathsf{action}_\alpha$},\textcolor{red}{$\mathsf{action}_\beta$}$,\ldots\in\mathit{Act}$, it is the monitor's turn to observe the environment, sets the turn flag to $t=3$ after each \textcolor{red}{monitor} action to trigger a controller \textcolor{red}{decision}, and restores the turn flag to $t=1$ immediately after that to enable the controlled system components to react to the new control parameters of the system. 
         }
         \label{subfig:generic-ideal-pDTMC}
     \end{subfigure}
     \hspace*{1.5mm}
     \begin{subfigure}[c]{0.547\linewidth}
         \centering
         \includegraphics[width=1.0\linewidth]{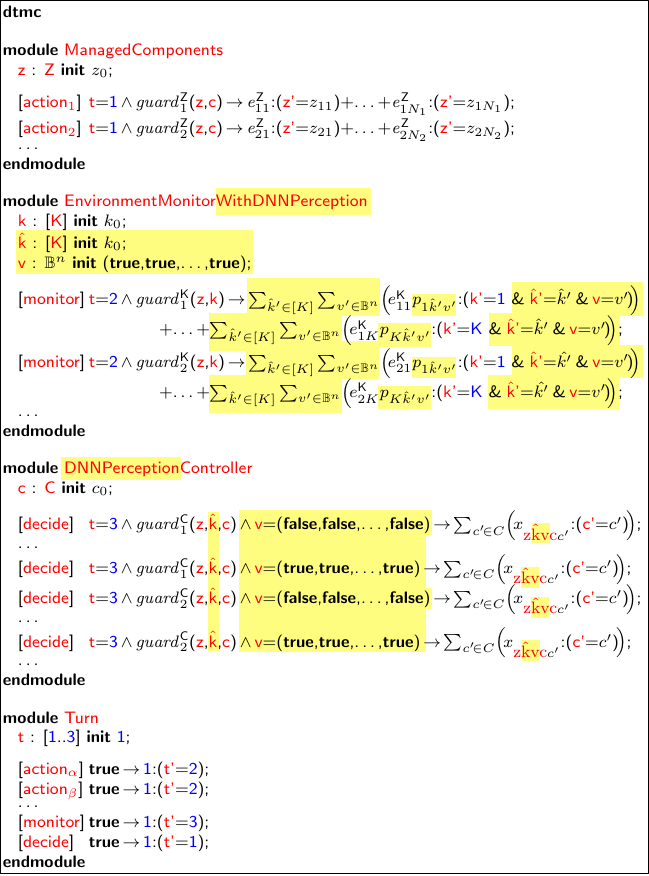}
         \caption{\acronym\ DNN-perception pDTMC model obtained by performing the highlighted modifications in the perfect-perception pDTMC from Figure~\ref{subfig:generic-ideal-pDTMC}. As a consequence of using a DNN to perceive the true environment state $k$, the \textcolor{red}{DNNPerceptionController} does not have access to its value; instead, it needs to rely on its estimate $\hat{k}$ and on the verification result $v$. The \textcolor{red}{EnvironmentMonitorWithDNNPerception} module continues to track the ground truth $k$ (necessary to establish the true safety, energy consumption, and other key properties of the system). Additionally, this module uses the DNN uncertainty quantification  probabilities~\eqref{eq:DNN-probabilities} to model the evolution of the DNN output $\hat{k}$ and the online DNN verification result $v$ associated with this output. The modules \textcolor{red}{ManagedComponents} and \textcolor{red}{Turn} are unchanged.}
         \label{subfig:generic-DNN-pDTMC}
     \end{subfigure}
     
     \caption{Perfect-perception and DNN-perception \acronym\ pDTMC models}
     \label{fig:model-augmentation}
\end{figure*}

%%%%%%%%%%%%%%%%%%%%%%%%%%%%
\medskip
\emph{b) Controller synthesis problem.} 
\acronym\ organises each state $s$ of the perfect-perception pDTMC model from Figure~\ref{fig:approach}a into a tuple 
\begin{equation}
  \label{eq:state-partition}
  s=(z,k,t,c),
\end{equation}
where $z\in Z$ models the (internal) state of the system, $k\in [K]$ models the state of the environment, $c\in C$ models the control parameters of the system, and $t\in [3]$ is a ``turn'' flag. This flag (i)~partitions the state set into states in which the system can change ($t=1$), states in which the environment is observed ($t=2$) and states in which it is the controller's ``turn'' to act ($t=3$); and (ii)~forces the pDTMC to visit these three types of states in this order:
\begin{equation}
\label{eq:pDTMC:properties}
\begin{array}{l}
  \forall s=(z,k,t,c), s'=(z',k',t',c')\in S :\\
    \quad \left((t=1 \;\wedge\; P(s,s')>0)  \implies k'=k \;\wedge\; c'=c \;\wedge\; t'<3\right) \;\wedge\\
    \quad \left((t=2 \;\wedge\; P(s,s')>0)  \implies z'=z \;\wedge\; c'=c \;\wedge\; t'=3 \right) \;\wedge\\
    \quad \left((t=3 \;\wedge\; P(s,s')>0)  \implies z'=z \;\wedge\; k'=k \;\wedge\; t'=1 \right).
\end{array}
\end{equation}
The outgoing transition probabilities from states $(z,k,3,c)\in S$ are controller parameters to be determined. We refer to them using the notation:
\begin{equation}
  \label{eq:ideal-controller-parameters}
    x_{zkcc'} = P((z,k,3,c),(z,k,1,c'))
\end{equation}
for all $c'\in C$, where $x_{zkcc'}\in\{0,1\}$ for \emph{deterministic controllers} or $x_{zkcc'}\in [0,1]$ for \emph{probabilistic controllers}, and $\sum_{c'\in C} x_{zkcc'} = 1$.

Figure~\ref{subfig:generic-ideal-pDTMC} shows the general format of a \acronym\ perfect-perception pDTMC model, defined in the high-level modelling language of the PRISM model checker.\cite{kwiatkowska_prism_2011}. In this language, a system is modelled by the parallel composition of a set of \emph{modules}. The state of a \emph{module} is given by a set of finite-range local variables, and its state transitions are specified by probabilistic guarded commands that change these variables:
\begin{equation}
  \label{eq:PrismCmd}
  [\mathit{action}]\;\;\;  \mathit{guard}\;  \rightarrow e_1:\mathit{update}_1 + e_2:\mathit{update}_2 + \ldots + e_m:\mathit{update}_N;
\end{equation}
where $\mathit{guard}$ is a boolean expression over the variables of all modules. If $\mathit{guard}$ evaluates to $\mathsf{true}$, the arithmetic expression $e_i$, $i\in [m]$, gives the probability with which the $\mathit{update}_i$ change of the module variables occurs. When $\mathit{action}$ is present, all modules comprising commands with 
this $\mathit{action}$ must \emph{synchronise} by performing one of these commands simultaneously. 

Given a pDTMC with these characteristics, the \emph{controller synthesis problem for the perfect-perception AS} is to find the combinations of parameter values for which the pDTMC satisfies $n_1 \geq 0$ PCTL-encoded \emph{constraints} 
\begin{equation}
  C_i::= \mathit{prop}_i\sim_i \mathit{bound}_i,
    \label{eq:constraints}
\end{equation}
and achieves optimal trade-offs among $n_2\geq 1$ PCTL-encoded \emph{objectives}  
\begin{equation}
  \hspace*{-3.5mm}O_j::= \textrm{minimise } \mathit{prop}_{n_1+j} \;\mid\; 
  \textrm{maximise } \mathit{prop}_{n_1+j}
   \label{eq:optimisation-objectives}
\end{equation}
where $\mathit{prop}_1$ to $\mathit{prop}_{n_1+n_2}$ are PCTL-encoded AS properties, $\sim_i\in\{<,\leq,\geq,>\}$, $\mathit{bound}_i\geq 0$, $i\in [n_1]$ and $j\in [n_2]$.

\medskip
\emph{c) pDTMC augmentation.} The controller of an AS with deep-learning perception cannot access the true environment state $k$ from~\eqref{eq:state-partition}. Instead, \acronym\ controllers need to operate with an estimate $\hat{k}\in [K]$ of $k$, and with the results $v=(v_1,v_2,\ldots,v_n)\in\mathbb{B}^n$ of the $n$ verification techniques~\eqref{eq:verif} applied to the DNN and its input that produced the estimate $\hat{k}$. The states $\hat{s}$ of a \acronym\ \emph{DNN-perception pDTMC model}, 
\begin{equation}
\label{eq:augmented-DTMC}
  \hat{\mathcal{M}}=(\hat{S},\hat{s_0},\hat{P},\hat{L},\hat{R}) 
\end{equation}
are tuples that extend~\eqref{eq:state-partition} with $\hat{k}$ and $v$:
\begin{equation}
  \label{eq:state-partition-with-DNN}
  \hat{s}=(z,k,\hat{k},v,t,c).
\end{equation}
The derivation of this pDTMC from the perfect-perception pDTMC is shown in Figure~\ref{subfig:generic-DNN-pDTMC}. 
To provide a formal definition of this derivation, we use the notation $s(\hat{s})=(z,k,t,c)$ to refer to the element from $Z\times [K]\times$ $[3]\times C$ that corresponds to a generic element $\hat{s}\in Z\times [K]^2\times\mathbb{B}^n\times[3]\times C$. With this notation, the elements of $\hat{\mathcal{M}}$ are obtained from the perfect-perception pDTMC $\mathcal{M}=(S, s_0, P, L, R)$ of the AS and the probabilities~\eqref{eq:DNN-probabilities} as follows:
\begin{equation}
\label{eq:augmented-DTMC-S}
\hat{S}=\{ \hat{s}\in Z\times [K]^2\times\mathbb{B}^n\times[3]\times C \mid s(\hat{s}) \in S\};
\end{equation}
\begin{equation}
\label{eq:augmented-DTMC-s0}
 \hat{s}_0=(z_0,k_0,k_0,\mathsf{true},\ldots,\mathsf{true},t_0,c_0), 
\end{equation}
where $(z_{0},k_{0},t_0,c_{0})=s_0$; and, for any states $\hat{s}=(z,k,\hat{k},v,t,c)$, $\hat{s}'=(z',k',\hat{k}',v',t',c')\in\hat{S}$,
\begin{equation}
\label{eq:augmented-DTMC-P}
\hat{P}(\hat{s},\hat{s}')= \left\{
 \begin{array}{ll}
 \hspace*{-1mm}P(s(\hat{s}),s(\hat{s}')), & \hspace*{-2mm}\textrm{if } t=1 \wedge (\hat{k}',v')=(\hat{k},v) \\[0.5mm]
 \hspace*{-1mm}P(s(\hat{s}),s(\hat{s}'))\cdot p_{k'\hat{k}'v'},
 & \hspace*{-2mm}\textrm{if } t=2   \\[0.5mm]
 \hspace*{-1mm}x_{z\hat{k}vcc'}, & \hspace*{-2mm}\textrm{if } t=3\; \wedge (z',k',\hat{k}',v',t')\\
&\qquad\qquad=(z,k,\hat{k},v,1)\\[0.5mm]
 \hspace*{-1mm}0, & \hspace*{-2mm}\textrm{otherwise}
 \end{array}
 \right.
\end{equation}
where 
\begin{equation}
    \label{eq:real-controller-parameters}
    x_{z\hat{k}vcc'} = \hat{P}((z,k,\hat{k},v,3,c), (z,k,\hat{k},v,1,c'))
\end{equation}
are controller parameters 
such that $x_{z\hat{k}vcc'}\in\{0,1\}$ for deterministic controllers or $x_{z\hat{k}vcc'}\in [0,1]$ for probabilistic controllers, and $\sum_{c'\in C}\; x_{z\hat{k}vcc'}=1$.
Finally, for any state $\hat{s}\in \hat{S}$,
\begin{equation}
\label{eq:augmented-DTMC-L}
 \hat{L}(\hat{s})=L(s(\hat{s})),
\end{equation}
and
\begin{multline}
\label{eq:augmented-DTMC-R}
 \hat{R} = \{(\hat{\rho},\hat{\iota}) \in (\hat{S}\rightarrow \mathbb{R}_{\geq 0}) \times (\hat{S}\times \hat{S}\rightarrow \mathbb{R}_{\geq 0})  \mid \\
 \qquad\qquad \exists (\rho,\iota)\in R : \bigl(\forall \hat{s}\in \hat{S} : \hat{\rho}(\hat{s})=\rho(s(\hat{s}))\bigr)\; \wedge\\[0.5mm]
 \qquad\qquad\qquad\qquad\;\;\;\: \bigl( \forall \hat{s},\hat{s}'\in \hat{S} : \hat{\iota}(\hat{s},\hat{s}') = \iota (s(\hat{s}), s(\hat{s}')) \bigr)\}
\end{multline}
An alternative encoding of the controller design space using a partially observable Markov decision process (POMDP) is possible.~\cite{7139019,chatterjee2016optimal} However, we opted for the pDTMC formalisation because current POMDP-enabled probabilistic model checkers~\cite{NPZ17,hensel2021probabilistic} do not support policy synthesis for combinations of requirements as complex as~\eqref{eq:constraints}, \eqref{eq:optimisation-objectives}.

The following result 
shows that the \acronym\ module augmentation produces a valid pDTMC in which the probabilities of control-parameter changes are independent of the true environment state $k$.

\begin{theorem}
\label{th:valid-pDTMC}
The tuple~\eqref{eq:augmented-DTMC} with the elements defined by~\eqref{eq:augmented-DTMC-S}--\eqref{eq:augmented-DTMC-R} is a valid pDTMC that satisfies:
\begin{equation}
\label{eq:DNN-perception-controller-independence-of-k}
\begin{array}{l}
  \forall \hat{s}=(z,k,\hat{k},v,t,c), \hat{s}'=(z',k',\hat{k}',v',t',c')\in \hat{S} :
   (c'\neq c \;\wedge\; \hat{P}(\hat{s},\hat{s}')>0)\\
   \quad\implies
   \bigl( (z',k',\hat{k}',v')=(z,k,\hat{k},v) \;\wedge\; t=3 \;\wedge\; t'=1 \;\wedge \\
   \qquad \forall k''\in[K]: \hat{P}((z,k'',\hat{k},v,t,c),(z,k'',\hat{k},v,t',c'))=\hat{P}(\hat{s},\hat{s}')\bigr).
\end{array}
\end{equation}
\end{theorem}

The next two theorems show that for each controller that satisfies constraints~\eqref{eq:constraints} and Pareto-optimises objectives~\eqref{eq:optimisation-objectives} for the DNN-perception AS there is an equivalent controller for the perfect-perception AS, but the converse does not hold.

\begin{theorem}
\label{th:DNN-to-perfect}
For any AS requirements \eqref{eq:constraints}, \eqref{eq:optimisation-objectives} for which there exists a DNN-perception controller that satisfies the constraints~\eqref{eq:constraints}, there exists also a perfect-perception controller that satisfies the same constraints and yields the same values for the PCTL properties from the optimisation objectives~\eqref{eq:optimisation-objectives}.
\end{theorem}

\begin{theorem}
\label{th:perfect-to-DNN}
If the confusion matrix $\mathcal{C}_{v_0}$ from~\eqref{eq:confusion} satisfies $\mathcal{C}_{v_0}[k_1,k_0]>0 \wedge \mathcal{C}_{v_0}[k_2,k_0]>0$ for a combination of verification results $v_0\in \mathbb{B}^n$, two classes $k_1\neq k_2$ and a class $k_0$, then there is an infinite number of AS requirements~\eqref{eq:constraints}, \eqref{eq:optimisation-objectives} for which there exists a  perfect-perception controller that satisfies the constraints~\eqref{eq:constraints}, and no DNN-perception controller exists that satisfies the constraints and yields the same values for the PCTL properties from the optimisation objectives~\eqref{eq:optimisation-objectives}.
\end{theorem}

Theorem~\ref{th:perfect-to-DNN} demonstrates that the decision-making capabilities of infinitely many perfect-perception controllers cannot be replicated by DNN-perception controllers (unless, exceptionally, applying the $n$ DNN verification techniques resolves the uncertainty introduces by the DNN). 
Finally, the following result shows that increasing the number of DNN verification techniques is never detrimental and may yield better \acronym\ controllers.

\begin{theorem}
\label{th:more-verif-methods}
For any AS requirements~\eqref{eq:constraints}, \eqref{eq:optimisation-objectives} and DNN-perception controller generated using $n$ DNN verification techniques $\mathit{verif}_1,\mathit{verif}_2,\ldots,\mathit{verif}_n$ such that the constraints~\eqref{eq:constraints} are satisfied, the \acronym\ pDTMC obtained using any verification technique $\mathit{verif}_{n+1}$ in addition to $\mathit{verif}_1,\mathit{verif}_2,\ldots,\mathit{verif}_n$ can be used to generate a controller that satisfies the constraints and yields the same values for the PCTL properties from the optimisation objectives.
\end{theorem}

\medskip
\noindent
\textbf{Stage 3: Controller synthesis} 

\medskip
\noindent
The \emph{controller synthesis problem for the DNN-perception system} involves finding instantiations for the controller parameters  for which the pDTMC $\hat{\mathcal{M}}$ from~\eqref{eq:augmented-DTMC} satisfies the constraints~\eqref{eq:constraints} and is Pareto optimal with respect to the optimisation objectives~\eqref{eq:optimisation-objectives}. Solving the general version of this problem precisely is unfeasible. However, metaheuristics such as multi-objective genetic algorithms for probabilistic model synthesis~\cite{calinescu_efficient_2018,gerasimou2018synthesis} can be used to generate close approximations of the Pareto-optimal controller set. Alternatively, exhaustive search can be employed to synthesise the actual Pareto-optimal controller set for AS with deterministic controllers and small numbers of parameters, or---by discretising the search space---an approximate Pareto-optimal controller set for AS with probabilistic controllers. We demonstrate the synthesis of \acronym~controllers through the use of both metaheuristics and exhaustive search in the next section.

%%%%%%%%%%%%%%%%%%%%%%%%%%%%%%%%%%%%%%%%%%%%%%%%%%%%%%%
\subsection*{2 \acronym~Applications}

We synthesised \acronym\ controllers for two simulated AS, with setups corresponding to the use of all possible combinations of two DNN verification techniques---minimum calibrated confidence threshold~\cite{GuoPSW17} ($\mathit{verif}_1$) and local robustness certification~\cite{leino21gloro} ($\mathit{verif}_2$)---in the \acronym\ DNN uncertainty quantification stage: (i)~no verification technique; (ii)~$\mathit{verif}_1$; (iii)~$\mathit{verif}_2$; and (iv)~$\mathit{verif}_1$ and $\mathit{verif}_2$. These \acronym\ applications are summarised below, with further details available in Methods, and the associated perfect-perception and DNN-perception pDTMCs described in the supplementary material and provided on our project website~\cite{DeepDECS-website}.

%%%%%%%%%%%%%%%%%%%%%%%%%%
\medskip
\noindent
\textbf{Mobile-robot collision limitation} 

\begin{figure*}
     \centering
     \includegraphics[width=0.63\linewidth]{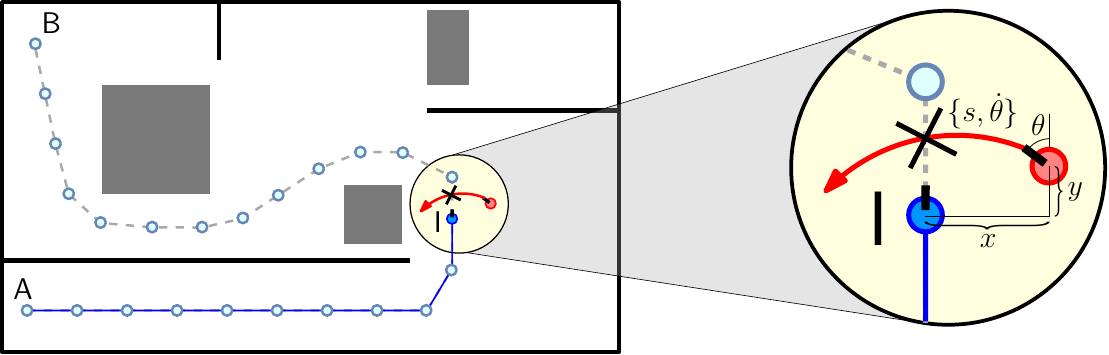}
     \caption{Collision limitation for a mobile robot tasked with traversing a known environment through the use of waypoints. 
     A mobile robot (darker blue) travelling between locations A and B 
     may collide with another mobile agent (red) when moving from its current waypoint I to the next. A two-class DNN predicts whether the robot is on collision course based on the relative horizontal distance $x$ and vertical distance $y$ between the robot and the collider, and the speed $s$, angle $\theta$ and angular velocity $\dot{\theta}$ of the collider.
        }
        \label{fig:collision-limitation}

        \vspace*{-3mm}
\end{figure*}

\medskip
\noindent
Inspired by recent research on DNN-based collision avoidance for autonomous aircraft~\cite{julian2019deep,julian2021reachability}, marine vehicles~\cite{xu2018deep} and  robots,~\cite{ehlers17} we used \acronym\ to develop a mobile robot collision-limitation controller. We considered a robot travelling between locations A and B, e.g., to carry goods in a warehouse (Figure~\ref{fig:collision-limitation}). Within this environment, the robot may encounter and potentially collide with another moving agent. We assume that collisions are not catastrophic, but should be limited to reduce robot damage and delays. As such, the robot uses DNN perception at each waypoint, to assess if it is on collision course. Based on the DNN output, its controller decides whether the robot will proceed to the next waypoint or should wait for a while at its current one.

\begin{figure*}
    \centering
    \begin{subfigure}[c]{\linewidth}
    \centering
    \includegraphics[width=1\textwidth]{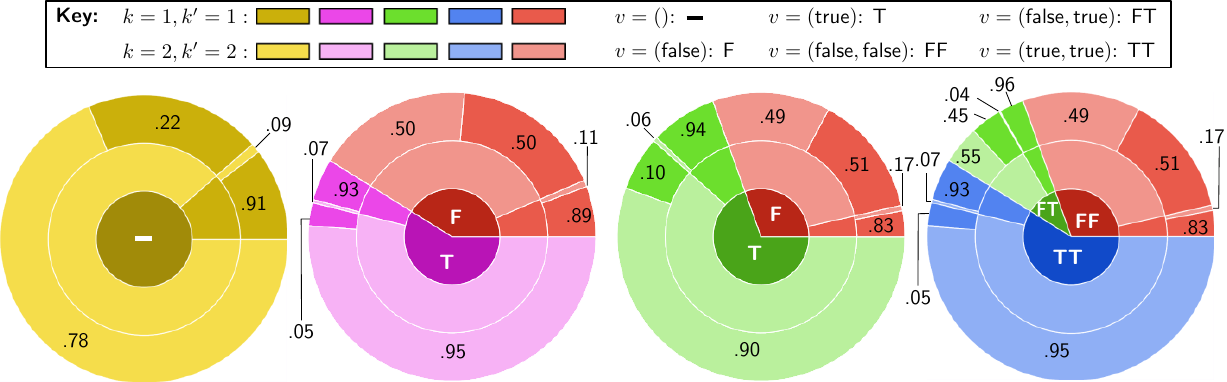}

\vspace*{-2mm}
\begin{flushleft}
\begin{small}
\hspace*{3mm}(i) no verification technique \hspace*{2.3cm}(ii) $\mathit{verif}_1$  \hspace*{3.2cm}(iii) $\mathit{verif}_2$ \hspace*{2.5cm}(iv) $\mathit{verif}_1$ and $\mathit{verif}_2$
\end{small}
\end{flushleft}

     \vspace*{-2mm}
     \caption{
     DNN uncertainty quantification results for the verification setups (i)--(iv). For every $v\in\mathbb{B}^n$ and $k,k'\in [K]$, the central disk shows the fraction of the test dataset included in the subset $X_v$ from~\eqref{eq:subset}; the inner ring shows the fraction of $X_v$ that belongs to class $k$; and the outer ring shows the probability that the DNN classifies an input of class $k$ from $X_v$ as class $k'$. We note that the correct classification probabilities (which correspond to identically shaded regions of the inner and outer rings) are much higher for verified DNN inputs (i.e., for $v=(\mathsf{true})$ in setups (ii) and (iii), and for $v\in\{(\mathsf{false},\mathsf{true}),(\mathsf{true},\mathsf{true})\}$ for setup (iv)) than for non-verified DNN inputs. Furthermore, the ``verified'' test subsets $X_{(\mathsf{true})}$ for setups (ii) and (iii) and $X_{(\mathsf{false},\mathsf{true})}\cup X_{(\mathsf{true},\mathsf{true})}$ for setup (iv) contain large fractions of the DNN test set; for setup (iv), all DNN inputs verified by $\mathit{verif}_2$ are also verified by $\mathit{verif}_1$, and therefore $X_{(\mathsf{true},\mathsf{false})}=\{\}$. 
     }
     \label{subfig:collision-limitation-confusion-matrices}
     \end{subfigure}
     
        \begin{subfigure}[b]{0.58\linewidth}
            \centering
            \includegraphics[width=1.0\textwidth,trim={0 0 0 7mm},clip]{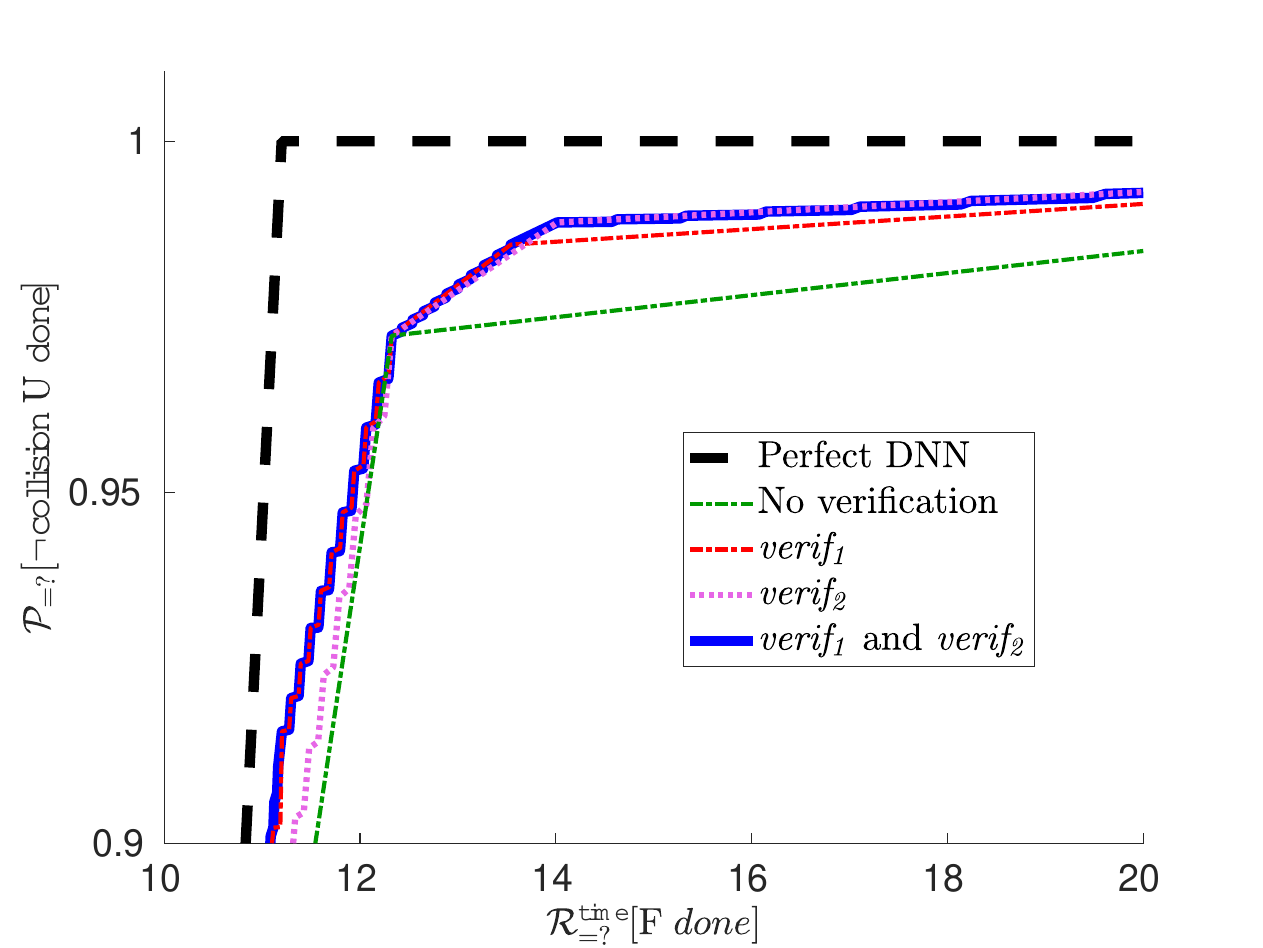}
            \caption{Pareto front associated with the set of Pareto-optimal \acronym\ controllers}    
            \label{subfig:collision-limitation-Pareto-fronts}
        \end{subfigure}
        \hfill
        \begin{subfigure}[b]{0.4\textwidth}  
            \centering 

\begin{tikzpicture}
\begin{axis}[
	xbar, xmin=0.1, xmax=0.22,
	bar width=8pt,
    width=5.8cm, height=3.2cm, enlarge y limits=0.15,
    y axis line style = { opacity = 0},
    symbolic y coords={{$verif_1$ and $verif_2$}, $verif_2$, $verif_1$, no verification},
    axis x line*       = bottom,
    tickwidth         = 0pt,
    tick label style={font=\footnotesize},
    ytick = data,
    nodes near coords, 
    every node near coord/.append style={font=\small},
    title  = IGD,
    title style={font=\small, at={(0.5,0.85)}}
]
\addplot coordinates {(0.11,{$verif_1$ and $verif_2$})  (0.1348,$verif_2$) (0.1426,$verif_1$) (0.2002,no verification)};
\end{axis}
\end{tikzpicture}
            
\vspace*{3mm}
\begin{tikzpicture}
\begin{axis}[
    xbar, xmin=9.8, xmax=10.06,
    bar width=8pt,
    width=5.8cm, height=3.2cm, enlarge y limits=0.15,
    y axis line style = { opacity = 0},
    symbolic y coords={{$verif_1$ and $verif_2$}, $verif_2$, $verif_1$, no verification},
    axis x line*       = bottom,
    tickwidth         = 0pt,
    tick label style={font=\footnotesize},
    ytick = data,
    nodes near coords, 
    every node near coord/.append style={font=\small},
    title  = HV,
    title style={font=\small, at={(0.5,0.85)}}
]
\addplot coordinates {(10.033826898252837,{$verif_1$ and $verif_2$})  (10.01019883419208,$verif_2$) (9.998695021953711,$verif_1$) (9.827539488225678,no verification)};
\end{axis}
\end{tikzpicture}
            
            \caption{Inverted Generational Distance (IGD) and hypervolume (HV) quality indicator values for the DNN-perception controller Pareto fronts: smaller IGD values and larger HV values indicate Pareto fronts that are better (i.e., closer to the ideal-perception Pareto front).}    
            \label{subfig:collision-limitation-Pareto-fronts-evaluation}
        \end{subfigure}
        
        \vskip\baselineskip
        \begin{subfigure}[b]{\textwidth}   
            \centering 
            \includegraphics[width=0.33\textwidth]{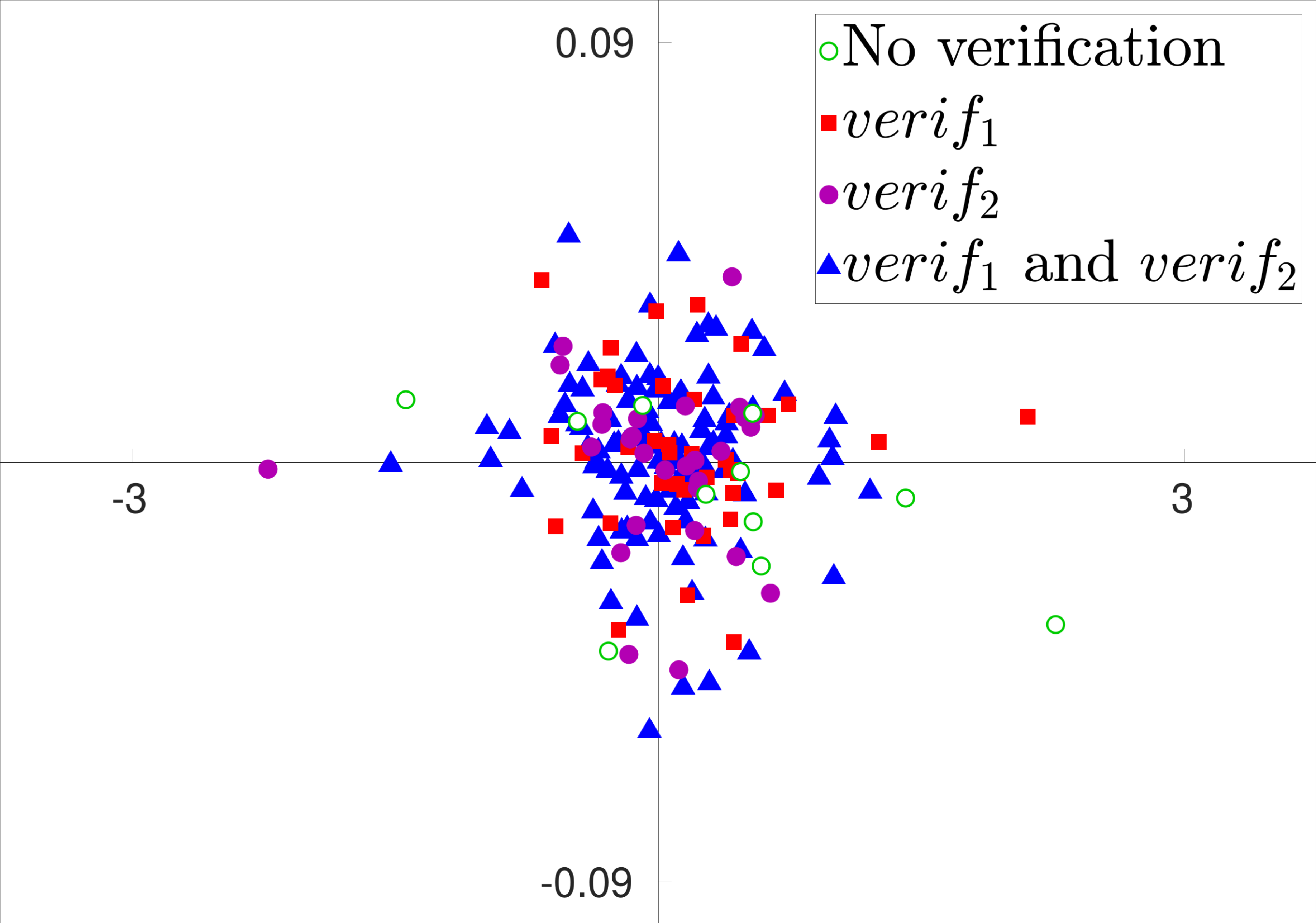}
            \includegraphics[width=0.33\textwidth]{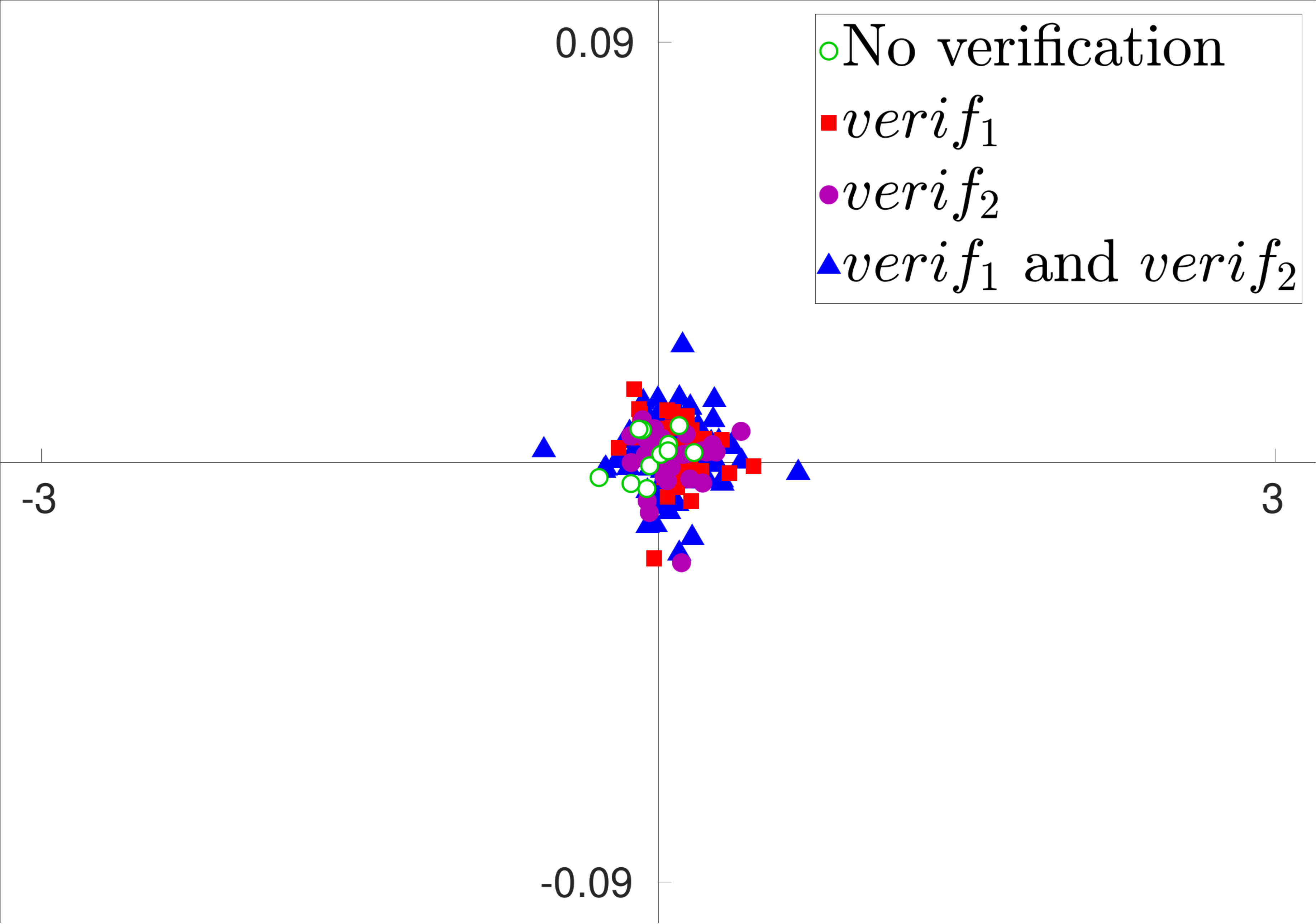}
            \includegraphics[width=0.33\textwidth]{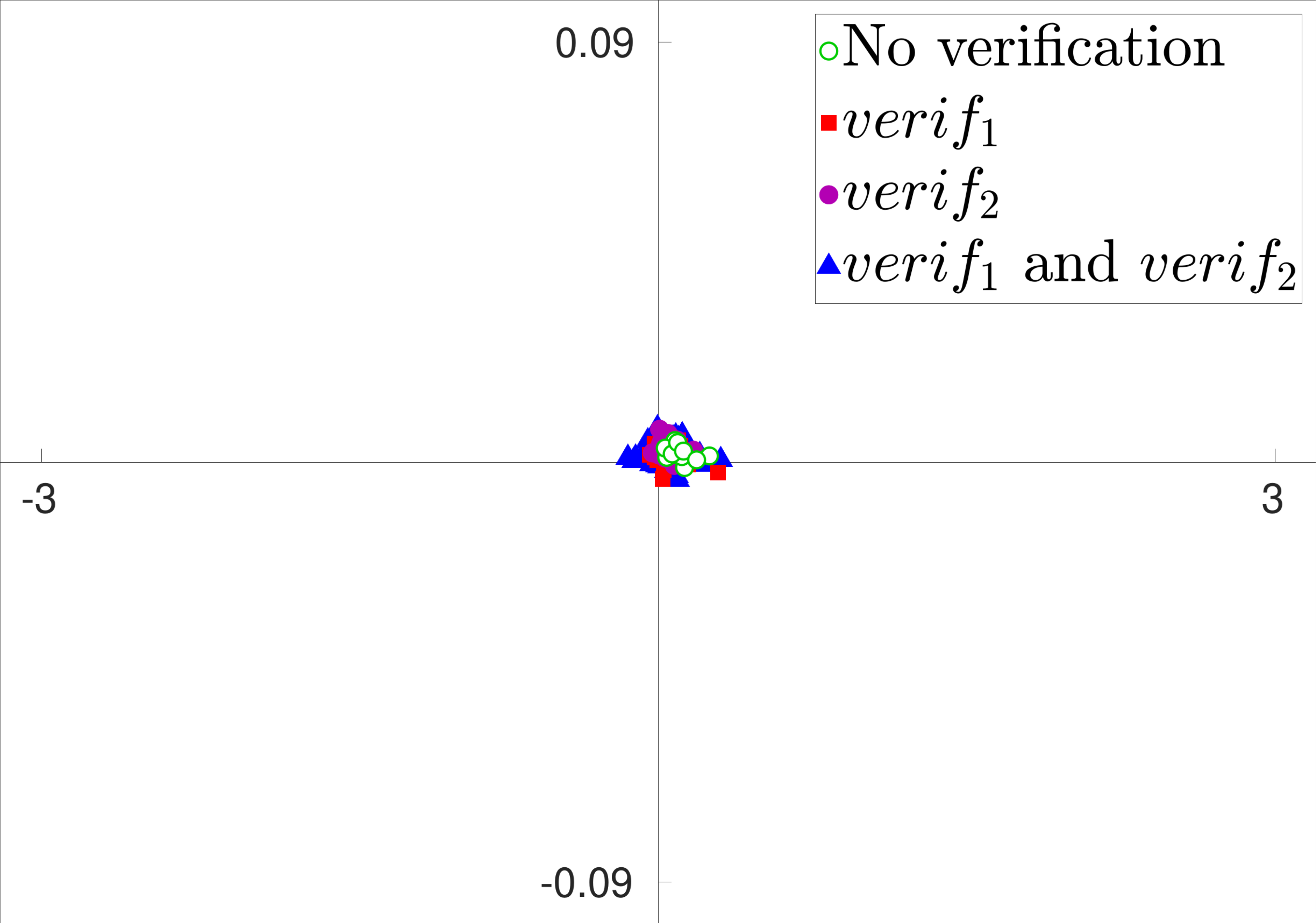}
            \caption{Difference between the model-predicted and experimental probability of collision-free travel and travel time for the Pareto-optimal \acronym\ controllers from Figure~\ref{subfig:collision-limitation-Pareto-fronts}, for journeys comprising  100~waypoints (left), 1000~waypoints (middle) and 10000~waypoints (right).}    
            \label{subfig:collision-limitation-testing}
        \end{subfigure}
        
        \caption{\acronym\ controller synthesis and testing results for the mobile robot collision limitation} 
        \label{fig:collision-limitation-results}
\end{figure*}

We trained a collision-prediction DNN using data from a simulator of the scenario in Figure~\ref{fig:collision-limitation}. We then applied \acronym\ to this DNN, a test dataset collected using our simulator, a perfect-perception pDTMC modelling the robot behaviour, and PCTL-encoded controller requirements demanding a collision-free journey with probability of at least $0.75$:
\begin{equation}
    C_1:\qquad\mathcal{P} [\neg \mathsf{collision}\; \mathrm{U}\; \mathsf{done}] \geq 0.75
    \label{eq:collision_constraint}
\end{equation}
and an optimal trade-off between maximising this probability and minimising the travel time:
\begin{equation}
    \begin{array}{ll}
    O_1: & \textrm{maximise }  \mathcal{P} [\neg \mathsf{collision}\; \mathrm{U}\; \mathsf{done}]\\
    O_2: & \textrm{minimise }  \mathcal{R}^\mathsf{time} [\mathrm{F}\; \mathsf{done}]
    \end{array}
    \label{eq:collision_objectives}
\end{equation}
The controller parameters synthesised by \acronym\ were the probabilities $x_{1v}$ and $x_{2v}$ for the robot to wait at its current waypoint when the DNN predicts it is not on collision course (class~1) and on collision course (class~2), respectively, where $v=()$ for setup (i), $v\in\mathbb{B}$ for setups (ii) and (iii), and $v\in\mathbb{B}^2$ for setup (iv) is the verification result for the DNN input that the prediction is based on. 

The \acronym\ results are presented in Figure~\ref{fig:collision-limitation-results}. The probabilities  
of the DNN classifying class-$k$ inputs associated with every verification result $v$ as class $k'$ are summarised in Figure~\ref{subfig:collision-limitation-confusion-matrices}, which shows that ``verified'' classifications (i.e., those associated with $v=(\mathsf{true})$ for setups (ii) and (iii), and $v=(\mathsf{true},\mathsf{true})$ for setup (iv)) are obtained for large percentages of DNN inputs, and have a much higher probability of being correct than ``unverified'' classifications. 

The controller design space was explored via discretising the controller parameters $x_{1v}, x_{2v}$, with each parameter varied between 0 and 1 with a step size of 0.1. The DNN-perception pDTMC instance for every parameter combination obtained in this way was analysed using PRISM. The Pareto fronts for the controllers that satisfied constraint~\eqref{eq:collision_constraint}  
for each setup are presented in Figure~\ref{subfig:collision-limitation-Pareto-fronts}, together with the Pareto front for the perfect-perception setup, which we analysed for comparison purposes. Expectedly, the best results are achieved in the perfect-perception setup, and the worst when no DNN verification technique is used. The use of verification methods yields Pareto fronts located closer to the perfect-perception Pareto front, with the best DNN-perception Pareto front obtained when both verification methods are used. These findings  
are confirmed by the analysis (Figure~\ref{subfig:collision-limitation-Pareto-fronts-evaluation}) of the Pareto fronts using two established Pareto-front quality indicators that we describe in Methods. 

To validate \acronym, we implemented and tested the Pareto-optimal controllers within our mobile-robot simulator. For each controller, we simulated 100~robot journeys comprising 100, 1000 and 10000 waypoints (i.e., 300~journeys in total). As shown in Figure~\ref{subfig:collision-limitation-testing}, the differences between the model-predicted and experimental journey time and probability of robot collision (averaged over the 100~journeys) decrease rapidly as the number of journey waypoints goes up, thus validating the \acronym\ models (whose analysis yields mean values for these AS properties) and controllers.

%%%%%%%%%%%%%%%%%%%%%%%%%%%%%%%
\bigskip
\noindent
\textbf{Driver-attentiveness management} 

\begin{figure*}
     \centering
     \includegraphics[width=0.46\linewidth]{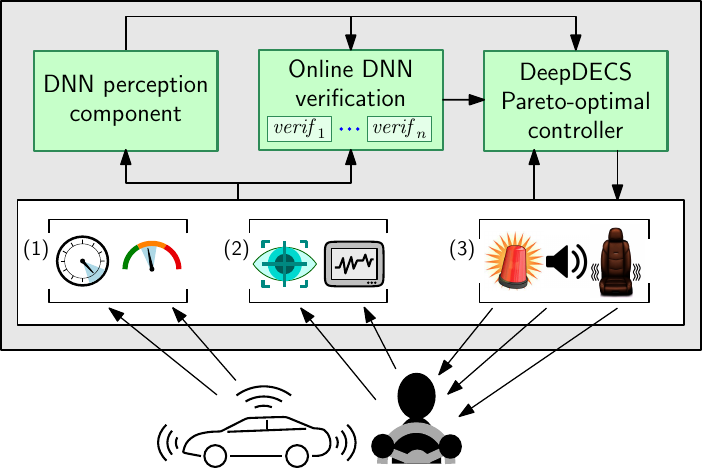}
        \caption{Driver-attentiveness management for shared-control autonomous driving. Data from car sensors (1) and driver biometric sensors (2) are supplied to a DNN perception component that classifies the driver state as attentive, semi-attentive or inattentive. The \acronym\ controller decides when optical, acoustic and/or haptic alerts (3) should be used to increase the driver's attentiveness.}
        \label{fig:SafeSCAD}

        \vspace*{-3mm}
\end{figure*}

\medskip
\noindent
We used \acronym\ to design a proof-of-concept driver-attentiveness management system for shared-control autonomous cars. Developed as part of our SafeSCAD project~\cite{SafeSCAD} and inspired by the first United Nations regulation on vehicles with Level~3 automation,~\cite{UNECE-2020} this system uses (Figure~\ref{fig:SafeSCAD}): (i)~specialised sensors to monitor key car parameters (velocity, lane position, etc.) and driver's biometrics (eye movement, heart rate, etc.), (ii)~a three-class DNN to predict the driver's response to a request to resume manual driving, and (iii)~a deterministic controller to issue visual/acoustic/haptic alerts when the driver is insufficiently attentive. 

\begin{figure*}

        \begin{subfigure}[b]{1.0\linewidth}
            \centering
                \includegraphics[width=1.0\textwidth]{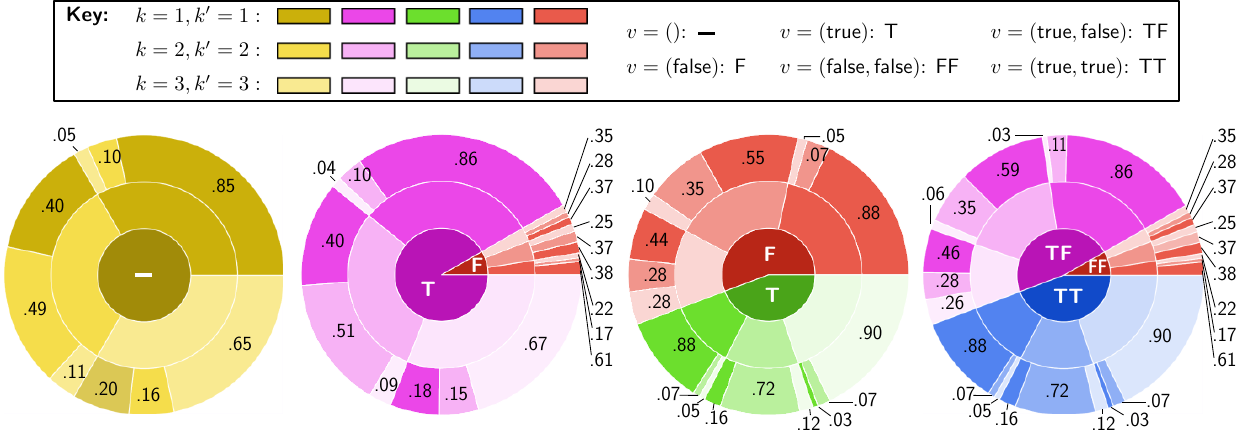}

\vspace*{-2mm}
\begin{flushleft}
\begin{small}
\hspace*{3mm}(i) no verification technique \hspace*{2.1cm}(ii) $\mathit{verif}_1$  \hspace*{3.1cm}(iii) $\mathit{verif}_2$ \hspace*{2.8cm}(iv) $\mathit{verif}_1$ and $\mathit{verif}_2$
\end{small}
\end{flushleft}
                
\vspace*{-2mm}
\caption{DNN uncertainty quantification results for the verification setups (i)--(iv). For every $v\in\mathbb{B}^n$ and $k,k'\in [K]$, the central disk shows the fraction of the test dataset included in the subset $X_v$ from~\eqref{eq:subset}; the inner ring shows the fraction of $X_v$ that belongs to class $k$; and the outer ring shows the probability that the DNN classifies an input of class $k$ from $X_v$ as class $k'$.
     We note that the ``verified'' test subsets $X_{(\mathsf{true})}$ for setups (ii) and (iii) and $X_{(\mathsf{true},\mathsf{false})}\cup X_{(\mathsf{true},\mathsf{true})}$ for setup (iv) contain large fractions of the DNN test dataset; for setup (iv), all DNN inputs verified by $\mathit{verif}_2$ are also verified by $\mathit{verif}_1$, and therefore $X_{(\mathsf{false},\mathsf{true})}=\{\}$. Additionally, the correct prediction probabilities (which correspond to identically shaded regions of the inner and outer rings) for the setups that use verification are higher for verified DNN inputs than for non-verified DNN inputs.}
     \label{subfig:SafeSCAD-confusion-matrices}
     \end{subfigure}

\vspace*{2mm}
        \begin{subfigure}[b]{0.6\linewidth}
            \centering
            \includegraphics[width=1.0\textwidth]{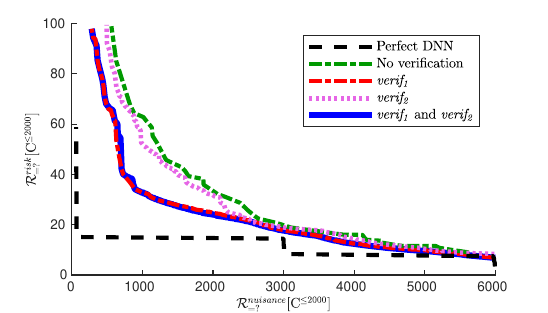}
            \caption{Pareto front associated with the set of Pareto-optimal SafeSCAD controllers} 
            \label{subfig:SafeSCAD-Pareto-fronts}
        \end{subfigure}
        \hfill
        \begin{subfigure}[b]{0.35\textwidth}  
            \centering 

            \begin{tikzpicture}
            \begin{axis}[
            	xbar, xmin=0., xmax=400,
            	bar width=8pt,
                width=5.8cm, height=3.2cm, enlarge y limits=0.15,
                y axis line style = { opacity = 0},
                symbolic y coords={{$verif_1$ and $verif_2$}, $verif_2$, $verif_1$, no verification},
                axis x line*       = bottom,
                tickwidth         = 0pt,
                tick label style={font=\footnotesize},
                ytick = data,
                nodes near coords, 
                every node near coord/.append style={font=\small},
                title  = IGD,
                title style={font=\small, at={(0.5,0.85)}}
            ]
            \addplot coordinates {(40.780590890969286,{$verif_1$ and $verif_2$})  (77.85462211181428,$verif_2$) (48.35022790259638,$verif_1$) (300.1610294763761,no verification)};
            \end{axis}
            \end{tikzpicture}

            \vspace*{3mm}
            \begin{tikzpicture}
            \begin{axis}[
                xbar, xmin=8, xmax=9.2,
                bar width=8pt,
                width=5.8cm, height=3.2cm, enlarge y limits=0.15,
                y axis line style = { opacity = 0},
                symbolic y coords={{$verif_1$ and $verif_2$}, $verif_2$, $verif_1$, no verification},
                axis x line*       = bottom,
                tickwidth         = 0pt,
                tick label style={font=\footnotesize},
                tick scale binop=\times,
                ytick = data,
                nodes near coords, 
                every node near coord/.append style={font=\small},
                title  = {HV [$\times 10^5$]},
                title style={font=\small, at={(0.5,0.85)}}
            ]
            \addplot coordinates {(8.93969,{$verif_1$ and $verif_2$})  (8.52000,$verif_2$) (8.92589,$verif_1$) (8.31493,no verification)};
            \end{axis}
            \end{tikzpicture}
            
            \caption{Evaluation of Pareto front quality using the established IGD and HV metrics.}    
            \label{subfig:SafeSCAD-Pareto-fronts-evaluation}
        \end{subfigure}

        \vspace*{1mm}
        \caption{\acronym\ controller synthesis results for the driver-attentiveness management system}
        \label{fig:SafeSCAD-results}

        \vspace*{2mm}
\end{figure*}

We used an existing DNN trained and validated with driver data from a SafeSCAD user study performed within a driving simulator,~\cite{pakdamanian2021deeptake} with the test dataset used for our DNN uncertainty quantification obtained from the same study. The controller requirements comprise two constraints that limit the maximum expected risk and driver nuisance cumulated over a 45-minute driving trip, and two optimisation objectives requiring that the same two measures are minimised: 
\begin{equation}
    \begin{array}{ll}
    C_1: & \mathcal{R}^\mathsf{risk} [\mathrm{C}^{\leq \mathit{ntrans}(45)}] \leq 100\\[1.5mm]
    C_2: & \mathcal{R}^\mathsf{nuisance} [\mathrm{C}^{\leq \mathit{ntrans}(45)}] \leq 6000\\[1.5mm]
    O_1: & \textrm{minimise }  \mathcal{R}^\mathsf{risk} [\mathrm{C}^{\leq \mathit{ntrans}(45)}]\\[1.5mm]
    O_2: & \textrm{minimise }  \mathcal{R}^\mathsf{nuisance} [\mathrm{C}^{\leq \mathit{ntrans}(45)}]
    \end{array}
    \label{eq:safescad-objectives}
\end{equation}
where $\mathcal{R}^\mathsf{rwd}[\mathrm{C}^{\leq \mathit{ntrans}(45)}]$ denotes the reward $\mathsf{rwd}$ cumulated over the number of DNN-perception pDTMC transitions corresponding to a 45-minute journey.
As the controller design space was too large for exhaustive exploration, we used the EvoChecker probabilistic model synthesis tool~\cite{gerasimou2018synthesis} to generate close approximations of the Pareto-optimal controllers. 

Figure~\ref{fig:SafeSCAD-results} shows the results of applying \acronym\ in this case study.
Similar to the robot collision-limitation controller, setups~(ii)--(iv) led to large fractions of the test dataset being verified, and to higher DNN accuracy for these subsets compared to the no-verification setup (Figure~\ref{subfig:SafeSCAD-confusion-matrices}), and the setups that employed verification techniques achieved Pareto-optimal controllers closer to the perfect-perception Pareto front (Figure~\ref{subfig:SafeSCAD-Pareto-fronts}). In particular, the knee points of the Pareto fronts from setups (ii) and (iv) are much closer to the knee point of the perfect-perception front than those from the other setups. These results are confirmed by the Pareto-front analysis (Figure~\ref{subfig:SafeSCAD-Pareto-fronts-evaluation}), which shows that the quality metrics for these two fronts are the best out of the four setups.

%%%%%%%%%%%%%%%%%%%%%%%%%%%%%%%%%%%%%%%%%%%%%%%%%
\vspace*{-3mm}
\subsection*{3 Discussion}

The use of deep-learning perception by autonomous systems poses a major challenge to traditional controller development methods. The \acronym\ method introduced in our paper addresses this challenge through several key contributions. 

First, it uses a suite of $n$ DNN verification techniques to identify, and to quantify the uncertainty of, $2^n$ categories of DNN outputs associated with different trustworthiness levels. This enables AS controllers to react confidently to highly trustworthy DNN outputs, and conservatively to untrustworthy ones by exploiting recent techniques for verifying DNN properties such as local robustness and confidence.\cite{leino21gloro, gopinath2018deepsafe, katz2019marabou, DBLP:conf/aitest/PatersonCP21, 10.1007/978-3-319-63387-9_1, singh19, GuoPSW17} 
Second, \acronym\ is underpinned by a new theoretical foundation for integrating DNN-perception uncertainty into discrete-time stochastic models of AS behaviour, enabling the formal analysis of safety, dependability and performance of AS with DNN perception. Third, it can use the models obtained in this way to synthesise both deterministic and probabilistic discrete-event controllers guaranteed to satisfy $n_1$ constraints, and Pareto-optimal with respect to $n_2$ optimisation objectives. 
Finally, it supports AS controller synthesis for different application domains, as shown by the case studies presented in the paper.

As \acronym\ is not prescriptive about the type of machine learning (ML) that introduces uncertainty into AS, we envisage that it is equally applicable to AS with other types of ML components for which local verification techniques exist to enable the quantification of their aleatory uncertainty. 
Such ML techniques that utilise confidence measures to quantify the uncertainty of their predictions include support vector machines and Gaussian processes.

The design of AS that use DNN classifiers for perception in combination with discrete-event controllers for decision-making has been studied before. The approach of Jha et al. \cite{jha-jar18} synthesises correct-by-construction controllers for AS with noisy sensors, i.e., with perception uncertainty. Unlike \acronym, this approach only considers systems that use linear models (i.e., not DNNs) for perception, and assumes already known uncertainty quantities. Moreover, while we formulate the control problem as a pDTMC, Jha et al.\ consider the simpler setting of deterministic linear systems.
Michelmore et al. \cite{michelmore20} analyze the safety of autonomous driving control systems that use DNNs in an end-to-end manner for both perception and control, i.e., the DNN consumes sensor readings and outputs control actions. They use Bayesian methods for calculating the uncertainty in the DNN control actions, and, when this uncertainty exceeds pre-determined thresholds, the system defaults to executing fail-safe actions. In contrast, we synthesise controllers that use the quantified uncertainty of DNN perception to select optimal yet safe actions.
Cleaveland et al. \cite{cleaveland2021monotonic} study the verification of AS with ML-based perception. They are interested in situations where the controller has already been constructed and the uncertainty in the perception outcomes is known, so the only goal is to verify if the AS satisfies a required probabilistic specification of safety.
In conclusion, synthesising safe and optimal controllers that account for the uncertainty in the DNN outcomes is a novel contribution of \acronym. Additionally, our DNN uncertainty quantification mechanism,  which uses the outcomes of off-the-shelf DNN verifiers in a black-box manner, is also new.

%%%%%%%%%%%%%%%%%%%%%%%%%%%%%%%%%%%%%%%%%%%%%%%%%%%%%%%
\begin{small}
\subsection*{\begin{small}4 Methods\end{small}}

\smallskip
\noindent
\textbf{Probabilistic computation tree logic}

\medskip
\noindent
The AS requirements for the synthesis of \acronym\ controllers are formally specified in probabilistic computation tree logic (PCTL) extended with rewards. 

\begin{definition}
\emph{State PCTL formulae} $\Phi$ and \emph{path PCTL formulae} $\Psi$ over an atomic proposition set $\mathit{AP}$, and \emph{PCTL reward formulae} $\Phi_R$ over a $\mathsf{rwd}$ reward structure~\eqref{eq:reward-structure} are defined by the grammar:
\begin{equation}
\label{eq:pctl}
\!\!\!\begin{array}{l}
    \Phi::=  true \;\vert\; \alpha \;\vert\; \Phi \wedge \Phi \;\vert\; \neg \Phi 
    \;\vert\; \mathcal{P} [\Psi] \sim p\\
    \Psi::= X \Phi \;\vert\; \Phi\; \mathrm{U}\; \Phi \;\vert\; \Phi\; \mathrm{U}^{\leq k}\, \Phi\\
    \Phi_R ::= \mathcal{R}^\mathsf{rwd} [\mathrm{C}^{\leq k}]\sim r \;\vert\; \mathcal{R}^\mathsf{rwd} [\mathrm{F}\; \Phi] \sim r
\end{array}
\end{equation}
where $\alpha \in AP$ is an atomic proposition, $\sim \in \{\geq, >, <, \leq\}$ is a relational operator, $p \in [0,1]$ is a probability bound, $r \in \mathbb{R}_0^+$ is a reward bound, and $k \in \mathbb{N}_{>0}$ is a timestep bound.
\end{definition}

\noindent
The PCTL semantics~\cite{Hansson1994,bianco1995model,andova2003discrete} is defined using a satisfaction \mbox{relation $\models$} over the states of a DTMC~\eqref{eq:DTMC}. Given a state $s$ of this DTMC $\mathcal{M}$, $s\models \Phi$ means `$\Phi$ holds in state $s$', and we have: always $s\models true$; $s \models \alpha$ iff $\alpha\in L(s)$; $s \models \neg \Phi$ iff $\neg (s\models \Phi)$; and $s\models \Phi_1 \wedge \Phi_2$ iff $s\models \Phi_1$ and $s\models \Phi_2$. 
 The \emph{time-bounded until formula} $\Phi_1\, \mathrm{U}^{\leq k}\, \Phi_2$ holds for a \emph{path} (i.e., sequence of DTMC states $s_0s_1s_2\ldots$ such that $P(s_i,s_{i+1})>0$ for all $i>0$) iff $\Phi_1$ holds in the first $i<k$ path states and $\Phi_2$ holds in the $(i+1)$-th path state; and the \emph{unbounded until formula} $\Phi_1\,\mathrm{U}\, \Phi_2$ removes the bound $k$ from the  time-bounded until formula. 
The \emph{next formula} $X \Phi$ holds if $\Phi$ is satisfied in the next state. 
The semantics of the probability $\mathcal{P}$ and reward $\mathcal{R}$ operators are defined as follows: 
$\mathcal{P} [\Psi] \sim p$ specifies that the probability that paths starting at state $s$ satisfy a path property $\Psi$ is ${\sim p}$;
$\mathcal{R}^\mathsf{rwd} [C^{\leq k}] \sim r$ holds if the expected cumulated reward up to time-step $k$ is $\sim r$; and 
$\mathcal{R}^\mathsf{rwd} [F \Phi] \sim r$ holds if the expected reward cumulated before reaching a state satisfying $\Phi$ is $\sim r$. 

Removing  $\sim\!p$ (or $\sim\!r$) from~\eqref{eq:pctl} specifies that the calculation of the probability (or reward) is required, e.g., see the optimisation objectives from~\eqref{eq:collision_objectives} and~\eqref{eq:safescad-objectives}. We use the shorthand notation $\mathit{pmc}(\Phi,\mathcal{M})$ and $\mathit{pmc}(\Phi_R,\mathcal{M})$ for these quantities computed (using probabilistic model checking) for the initial state $s_0$ of $\mathcal{M}$.

\medskip
\noindent
\textbf{Theorem proofs}

\medskip
\noindent
\hspace*{-1mm}\renewenvironment{proof}{\medskip\noindent{\bfseries Proof of Theorem~\ref{th:valid-pDTMC}.}}{\qed\smallskip}
\begin{proof}
To demonstrate that~\eqref{eq:augmented-DTMC} is a valid pDTMC, we need to show that, for any state $\hat{s}=(z,k,\hat{k},v,t,c)\in\hat{S}$, $\sum_{\hat{s}'\in\hat{S}} \hat{P}(\hat{s},\hat{s}')=1$. We prove this and the following variant of~\eqref{eq:pDTMC:properties} (which is required for the subsequent proofs) 
\begin{equation}
\label{eq:augmentedpDTMC:properties}
\hspace*{-3mm}\begin{array}{l}
  \forall \hat{s}=(z,k,\hat{k},v,t,c), \hat{s}'=(z',k',\hat{k}',v',t',c')\in \hat{S} :\\
    \quad \bigl((t=1 \;\wedge\; P(\hat{s},\hat{s}')>0)  \implies (k',\hat{k}',v',c')=(k,\hat{k},v,c) \;\wedge\; t'<3\bigr) \;\wedge\\[1mm]
    \quad \bigl((t=2 \;\wedge\; P(\hat{s},\hat{s}')>0)  \implies (z',c')=(z,c) \;\wedge\; t'=3 \bigr) \;\wedge\\[1mm]
    \quad \bigl((t=3 \;\wedge\; P(\hat{s},\hat{s}')>0)  \implies (z',k',\hat{k}',v')=(z,k,\hat{k},v) \;\wedge\; t'=1 \bigr)
\end{array}
\end{equation}
for each possible value of $t$, i.e., for $t\in\{1,2,3\}$. 

For $t=1$, \eqref{eq:augmented-DTMC-P} implies that 
\[
  \sum_{\hat{s}'\in\hat{S}} \hat{P}(\hat{s},\hat{s}') =
  \sum_{\hat{s}'\in\hat{S}} P(s(\hat{s}),s(\hat{s}')) = 1  
\]
because the last sum adds up all outgoing transition probabilities of state $s(\hat{s})$ %$(z',k',t',c')$ 
from the perfect-perception pDTMC $\mathcal{M}$. Consider now any $\hat{s}'=(z',k',\hat{k}',v',t',c')\in \hat{S}$ such that $\hat{P}(\hat{s},\hat{s}')>0$. According to~\eqref{eq:augmented-DTMC-P}, this requires $\hat{k}'=\hat{k}\;\wedge\; v'=v$ when $t=1$. Additionally, since $\hat{P}(\hat{s},\hat{s}')=P((z,k,1,c),(z',k',t',c'))$, \eqref{eq:pDTMC:properties} implies that $k'=k\;\wedge\; c'=c\;\wedge\; t'<3$, as required by~\eqref{eq:augmentedpDTMC:properties}.

For $t=2$, we have
\begin{multline*}
  \sum_{\hat{s}'\in\hat{S}} \hat{P}(\hat{s},\hat{s}') =  \sum_{(z',k',\hat{k}',v',t',c')\in\hat{S}} \left(\hat{P}(\hat{s},(z',k',\hat{k}',v',t',c'))\cdot p_{k'\hat{k}'v'}\right) \\
     = \sum_{(z',k',t',c')\in S} \biggl( P((z,k,2,c),(z',k',t',c')) \cdot \sum_{(\hat{k}',v')'\in[K]\times\mathbb{B}^n} p_{k'\hat{k}'v'}\biggr) \\
     =  \sum_{(z',k',t',c')\in S} \left( P((z,k,2,c),(z',k',t',c')) \cdot 1 \right) = 1.
\end{multline*}
Consider again a generic $\hat{s}'=(z',k',\hat{k}',v',t',c')\in \hat{S}$ such that $\hat{P}(\hat{s},\hat{s}')>0$.  Since $\hat{P}(\hat{s},\hat{s}') = P((z,k,2,c),(z',k',t',c'))\cdot p_{k'\hat{k}'v'}$, \eqref{eq:pDTMC:properties} implies that $(z',c')=(z,c) \;\wedge\; t'=3$.

Finally, for $t=3$, we have $\sum_{\hat{s}'\in\hat{S}} \hat{P}(\hat{s},\hat{s}')\!=\!\sum_{c'\in C}\; x_{z\hat{k}vcc'}\!=\!1$ and the property~\eqref{eq:augmentedpDTMC:properties} is explicitly stated in~\eqref{eq:augmented-DTMC-P}.

To show now that~\eqref{eq:DNN-perception-controller-independence-of-k} holds, we note that, according to definition~\eqref{eq:augmented-DTMC-P}, both transition probabilities from this relation (i.e., $\hat{P}(\hat{s},\hat{s}')$ and $\hat{P}((z,k'',\hat{k},v,t,c),(z,k'',\hat{k},v,t',c'))$) are equal to $x_{z\hat{k}vcc'}$.
\end{proof}

\medskip
\noindent
To prove Theorems~\ref{th:DNN-to-perfect} and~\ref{th:perfect-to-DNN}, we require the following auxiliary result.

\begin{lemma}
\label{lemm:path-equivalence}
Let $\underline{x}$ and $\underline{\hat{x}}$ be valid instantiations of the perfect-perception controller parameters $\bigl\{x_{zkcc'}\in[0,1] \;\bigl|\; (\exists k\in[K].(z,k,3,c)\in S) \;\wedge\; c'\in C \bigr\}$ from~\eqref{eq:ideal-controller-parameters} and of the DNN-perception controller parameters $\bigl\{x_{z\hat{k}vcc'}\in[0,1] \;\bigl|\;(\exists k\in[K].(z,k,\hat{k},v,3,c)\in \hat{S}) \;\wedge\; c'\in C \bigr\}$ from~\eqref{eq:augmented-DTMC-P}, respectively. Also, let $\mathcal{M}_{\underline{x}}$ and $\hat{\mathcal{M}}_{\underline{\hat{x}}}$ be the instances of the perfect-per\-ception pDTMC $\mathcal{M}$ and DNN-perception pDTMC $\hat{\mathcal{M}}$ corresponding to the controller parameters $\underline{x}$ and $\underline{\hat{x}}$, respectively. With this notation, we have 
\begin{equation}
  \label{eq:equivalence}
  \mathit{pmc}(\Phi,\hat{\mathcal{M}}_{\underline{\hat{x}}}) = \mathit{pmc}(\Phi,\mathcal{M}_{\underline{x}}),
\end{equation}
and
\begin{equation}
  \label{eq:equivalence-R}
  \mathit{pmc}(\Phi_R,\hat{\mathcal{M}}_{\underline{\hat{x}}}) = \mathit{pmc}(\Phi_R,\mathcal{M}_{\underline{x}}),
\end{equation}
for any (quantitative) PCTL state formula $\Phi$ and reward state formula $\Phi_R$ if and only if the elements of $\underline{x}$ and $\underline{\hat{x}}$ satisfy
\begin{equation}
   \label{eq:equivalence-x}
   x_{zkcc'} = \sum_{\hat{k}\in[K]}\; \sum_{v\in \mathbb{B}^n} p_{k\hat{k}v}x_{z\hat{k}vcc'}
\end{equation}
for all $(z,k,3,c)\in S$ and $c'\in C$.
\end{lemma}
\renewenvironment{proof}{\medskip\noindent{\bfseries Proof.}}{\qed\smallskip}
\begin{proof}
Let $\mathit{Paths}^{\mathcal{M}_{\underline{x}}}(s_0)$ and $\mathit{Paths}^{\hat{\mathcal{M}}_{\underline{\hat{x}}}}(\hat{s}_0)$ be the set of all $\mathcal{M}_{\underline{x}}$ paths starting at $s_0$ and the set of all $\hat{\mathcal{M}}_{\underline{\hat{x}}}$ paths starting at $\hat{s}_0$, respectively. Equalities~\eqref{eq:equivalence} and~\eqref{eq:equivalence-R} hold iff, for any path $\pi=s_0s_1s_2\ldots\in \mathit{Paths}^{\mathcal{M}_{\underline{x}}}(s_0)$, set of associated paths $\hat{\Pi} =\bigl\{\hat{s}_0\hat{s}_1\hat{s}_2\ldots\in\mathit{Paths}^{\hat{\mathcal{M}}_{\underline{\hat{x}}}}(\hat{s}_0) \mid \forall i\geq 0\;.\; s(\hat{s}_i)=s_i \bigr\}$, and $i\geq 0$, the following property holds:
\begin{equation}
  \label{eq:equivalence-intermediate}
    P(s_i,s_{i+1}) = \sum_{\hat{s}_0\hat{s}_1\hat{s}_2\ldots\in\hat{\Pi}} \hat{P}(\hat{s}_i,\hat{s}_{i+1}).
\end{equation}
This is required because, according to~\eqref{eq:augmented-DTMC-L} and~\eqref{eq:augmented-DTMC-R}, the $(i+1)$-th state of $\pi$ and of any path $\hat{\pi}\in\hat{\Pi}$ are labelled with the same atomic propositions and assigned the same state rewards, respectively; and, according to~\eqref{eq:augmented-DTMC-R}, the transition rewards for the transition between their $i$-th state and $(i+1)$-th state are also identical. Thus, if this equality holds, the path $\pi$ and path set $\hat{\Pi}$ are indistinguishable in the evaluation of PCTL state and state reward formulae; and, if the equality does not hold, a labelling function $L$ and a PCTL state formula $\Phi$ (or state reward formula $\Phi_R$) can be handcrafted to provide a counterexample for~\eqref{eq:equivalence} (or for \eqref{eq:equivalence-R}). 

Given the definition of $\hat{P}$ from~\eqref{eq:augmented-DTMC-P}, property~\eqref{eq:equivalence-intermediate} holds trivially for any state $s_i=(z,k,t,c)\in S$ with $t=1$, and also holds for states $s_i$ with $t=2$ because 
\begin{multline*}
\sum_{\hat{s}_0\hat{s}_1\hat{s}_2\ldots\in\hat{\Pi}} \hat{P}(\hat{s}_i,\hat{s}_{i+1}) = \sum_{\hat{s}_0\hat{s}_1\hat{s}_2\ldots\in\hat{\Pi}} (P(s_i,s_{i+1})\cdot p_{k\hat{k}_{i+1}v_{i+1}})\\
=  P(s_i,s_{i+1}) \cdot \sum_{\hat{s}_0\hat{s}_1\hat{s}_2\ldots\in\hat{\Pi}} p_{k\hat{k}_{i+1}v_{i+1}} = P(s_i,s_{i+1})\cdot 1 = P(s_i,s_{i+1}),
\end{multline*}
where $\hat{k}_{i+1}$ and $v_{i+1}$ represent the DNN prediction and verification result for each state $\hat{s}_{i+1}$ from the sum, respectively. Finally, for $t=3$, property~\eqref{eq:equivalence-intermediate} holds if and only if the perfect-perception and DNN-perception controllers select each next controller configuration $c'\in C$ with the same probability for $s_i$ and for all the states $\hat{s}_i$ from $\hat{\Pi}$ taken together, i.e., if and only if~\eqref{eq:equivalence-x} holds, which completes the proof. \end{proof}

\renewenvironment{proof}{\medskip\noindent{\bfseries Proof of Theorem~\ref{th:DNN-to-perfect}.}}{\qed\smallskip}
\begin{proof}
We prove this result by showing that the application of~\eqref{eq:equivalence-x} to any valid instantiation of the DNN-perception controller parameters $x_{z\hat{k}vcc'}$ produces a valid instantiation of the perfect-perception controller parameters $x_{zkcc'}$. First, since $x_{z\hat{k}vcc'}\in[0,1]$ for any valid $(z,\hat{k},v,c,c')$ tuple, we have
\[
    0 =\hspace*{-1mm} \sum_{\hat{k}\in[K]}\; \sum_{v\in \mathbb{B}^n} (p_{k\hat{k}v} \cdot 0) \leq\hspace*{-1mm} \sum_{\hat{k}\in[K]}\; \sum_{v\in \mathbb{B}^n} p_{k\hat{k}v}x_{z\hat{k}vcc'} \leq\hspace*{-1mm}  \sum_{v\in \mathbb{B}^n} (p_{k\hat{k}v} \cdot 1) \leq 1, 
\]
so $x_{zkcc'}\in[0,1]$ for any valid tuple $(z,k,c,c')$. Additionally, for any valid combination of $z$, $k$ and $c$, we have
\begin{multline*}
\sum_{c'\in C} x_{zkcc'} = \sum_{c'\in C} \sum_{\hat{k}\in[K]}\; \sum_{v\in \mathbb{B}^n} p_{k\hat{k}v}x_{z\hat{k}vcc'} \\
 =  \sum_{c'\in C}  \left(x_{z\hat{k}vcc'} \cdot \left( \sum_{\hat{k}\in[K]}\; \sum_{v\in \mathbb{B}^n} p_{k\hat{k}v} \right)  \right) =  \sum_{c'\in C}  \left(x_{z\hat{k}vcc'} \cdot 1 \right) = 1,
\end{multline*}
which completes the proof.
\end{proof}

\renewenvironment{proof}{\medskip\noindent{\bfseries Proof of Theorem~\ref{th:perfect-to-DNN}.}}{\qed\smallskip}
\begin{proof}
Consider two perfect-perception controller parameters $x_{zk_1cc'}$ and $x_{zk_2cc'}$ corresponding to a configuration $c'$ being selected by the controller when the environment state is $k_1$ and $k_2$, respectively. 
Since $\mathcal{C}_{v_0}[k_1,k_0]>0 \wedge \mathcal{C}_{v_0}[k_2,k_0]>0$, definition~\eqref{eq:DNN-probabilities} implies that $p_{k_1k_0v_0}>0 \wedge p_{k_2k_0v_0}>0$, and we consider the infinite set of perfect-perception controllers with $x_{zk_1cc'}=1$ and $x_{zk_2cc'}\in[0,p_{k_2k_0v_0})$. For any such controller, consider the instantiation of equality~\eqref{eq:equivalence-x} for $x_{zk_1cc'}$. The parameters of any equivalent DNN-perception controller that appear on the right-hand side of this instantiation and are multiplied by non-zero probabilities $p_{k\hat{k}v}$ must have value $1$, or otherwise the terms of the double sum from~\eqref{eq:equivalence-x} will add up to a value below $1$. In particular, we must have $x_{zk_0v_0cc'}=1$ because this parameter is multiplied by $p_{k_1k_0v_0}>0$. However, according to~\eqref{eq:equivalence-x}, this means that the DNN-perception controller can only be equivalent to a perfect-perception controller whose parameter $x_{zk_2cc'}$ satisfies
\begin{multline*}
   x_{zk_2cc'} = \sum_{\hat{k}\in[K]}\; \sum_{v\in \mathbb{B}^n} p_{k_2\hat{k}v}x_{z\hat{k}vcc'}
   \geq p_{k_2k_0v_0}\cdot x_{zk_0v_0cc'}\\
   = p_{k_2k_0v_0}\cdot 1 = p_{k_2k_0v_0}.
\end{multline*}
This inequality is not satisfied by any of the perfect-perception controllers from the infinite set we considered. As such, no equivalent DNN-perception controller exists for any of these perfect-perception controllers, which completes the proof.
\end{proof}

\renewenvironment{proof}{\medskip\noindent{\bfseries Proof of Theorem~\ref{th:more-verif-methods}.}}{\qed\smallskip}
\begin{proof}
\mbox{Let $\hat{\mathcal{M}}^n$ and $\hat{\mathcal{M}}^{n+1}$ be the DNN-perception} pDTMCs obtained using the DNN verification techniques $\mathit{verif}_1,\mathit{verif}_2,\ldots,\mathit{verif}_n$, and the DNN verification techniques $\mathit{verif}_1,\mathit{verif}_2,\ldots,\mathit{verif}_{n+1}$, respectively. Consider any instantiation $\underline{\hat{x}}^n$ of the controller parameters~\eqref{eq:augmented-DTMC-P} for $\hat{\mathcal{M}}^n$ so that the constraints~\eqref{eq:constraints} are satisfied, and let $\underline{\hat{x}}^{n+1}$ be the instantiation of the controller parameters~\eqref{eq:augmented-DTMC-P} for $\hat{\mathcal{M}}^{n+1}$ such that the elements of this instantiation satisfy
\begin{equation}
\label{eq:more-verif-methods}
    \hspace*{-2.5mm}\underline{x}^{n+1}_{z\hat{k}(v_1,v_2,\ldots,v_n,\mathsf{false})cc'} =  \underline{x}^{n+1}_{z\hat{k}(v_1,v_2,\ldots,v_n,\mathsf{true})cc'} =  \underline{x}^{n}_{z\hat{k}(v_1,v_2,\ldots,v_n)cc'}.
\end{equation}
We will show that the controller defined by $\underline{\hat{x}}^{n+1}$ over $\hat{\mathcal{M}}^{n+1}$ is equivalent to the controller defined by $\underline{\hat{x}}^{n}$ over $\hat{\mathcal{M}}^{n}$ (and therefore must satisfy the constraints~\eqref{eq:constraints} and yield the same values for the PCTL properties from the optimisation objectives~\eqref{eq:optimisation-objectives}). 

According to Lemma~\ref{lemm:path-equivalence}, the former controller is equivalent to the perfect-perception controller whose parameters~\eqref{eq:ideal-controller-parameters} satisfy:
\begin{equation}
   \label{eq:equivalent-perfect-perception-controller}
   x_{zkcc'} = \sum_{\hat{k}\in[K]}\;\;\; \sum_{v\in \mathbb{B}^{n+1}} p_{k\hat{k}v}\underline{x}^{n+1}_{z\hat{k}vcc'}.
\end{equation}
Taking into account~\eqref{eq:more-verif-methods}, we obtain:
\begin{multline*}
   \hspace*{-3mm}x_{zkcc'} = \sum_{\hat{k}\in[K]}\;\;\; \sum_{(v_1,v_2,\ldots,v_{n})\in \mathbb{B}^{n}} \left( p_{k\hat{k}(v_1,v_2,\ldots,v_n,\mathsf{false})}\underline{x}^{n+1}_{z\hat{k}(v_1,v_2,\ldots,v_n,\mathsf{false})cc'}\right.\\
   +\left. p_{k\hat{k}(v_1,v_2,\ldots,v_n,\mathsf{true})}\underline{x}^{n+1}_{z\hat{k}(v_1,v_2,\ldots,v_n,\mathsf{true})cc'} \right)\\
   = \sum_{\hat{k}\in[K]}\;\;\; \sum_{(v_1,v_2,\ldots,v_{n})\in \mathbb{B}^{n}} \left[ \left( p_{k\hat{k}(v_1,v_2,\ldots,v_n,\mathsf{false})} + p_{k\hat{k}(v_1,v_2,\ldots,v_n,\mathsf{true})}\right)\right.\\
    \left. \cdot \underline{x}^{n}_{z\hat{k}(v_1,v_2,\ldots,v_n)cc'}\right] 
   = \sum_{\hat{k}\in[K]}\;\;\; \sum_{v\in \mathbb{B}^{n}} p_{k\hat{k}v}\underline{x}^n_{z\hat{k}vcc'}
\end{multline*}
because, according to~\eqref{eq:DNN-probabilities}, 
\begin{multline*}
p_{k\hat{k}(v_1,v_2,\ldots,v_n,\mathsf{false})} + p_{k\hat{k}(v_1,v_2,\ldots,v_n,\mathsf{true})} \\
= \frac{C_{(v_1,v_2,\ldots,v_n,\mathsf{false})}[k,k'] + C_{(v_1,v_2,\ldots,v_n,\mathsf{true})}[k,k']}{\sum_{v'\in\mathbb{B}^{n+1}} \sum_{k''\in[K]} C_{v'}[k,k'']}\\
= \frac{C_{(v_1,v_2,\ldots,v_n)}[k,k']}{\sum_{v'\in\mathbb{B}^n} \sum_{k''\in[K]} C_{v'}[k,k'']} = p_{k\hat{k}(v_1,v_2,\ldots,v_n)}.
\end{multline*}
According again to Lemma~\ref{lemm:path-equivalence}, this result implies that the perfect-perception controller induced by~\eqref{eq:equivalent-perfect-perception-controller} is also equivalent to the controller defined by $\underline{\hat{x}}^{n}$ over $\hat{\mathcal{M}}^{n}$. Hence, using~\eqref{eq:more-verif-methods} to select the parameters of the controller obtained for $n+1$ DNN verification technique yields a controller equivalent to that obtained using only the first $n$ verification techniques. 
\end{proof}

\bigskip
\noindent
\textbf{Robot collision limitation}

\medskip
\noindent
\textbf{Dataset.} The dataset for training the DNN and quantifying its uncertainty was obtained using the 2D particle simulator \textit{Box2D} (\url{https://box2d.org/}), with the robot and collider simulated by circular particles of 0.5-unit radius. We ran simulations with the robot starting at the origin $(0,0)$ with a heading of $\frac{\pi}{2}$ radians, and travelling in a straight line to a goal destination $(x_\mathit{goal}, y_\mathit{goal})$, with a speed of 1~unit/s. A journey was deemed completed when the robot reached a goal area defined by $(x_\mathit{goal}\pm \epsilon, y_\mathit{goal}\pm \epsilon) $ for a small $\epsilon>0$. The robot advanced with an angular velocity
\begin{equation*}
    \dot{\theta}_{r}=\alpha \cdot \arctan \left( \frac{v_{x}\cdot y_{goal} - v_{y}\cdot x_{goal}}{v_{x}\cdot x_{goal}+v_{y}\cdot y_{goal}} \right)
\end{equation*}
where $v_{x}$ and $v_{y}$ are the horizontal and vertical velocities of the robot, respectively, and $\alpha>0$ is a constant. When the difference between the robot's and the target heading exceeded $\frac{\pi}{36}$, the robot's linear speed was reduced to 0.1~unit/s, allowing it time to correct its course. 
The collider had a random initial position 
\begin{equation*}
    (x,\; y,\; \theta) = (\mathcal{U}(-x_{lim}, x_{lim}),\; \mathcal{U}(0, y_{lim}),\; \mathcal{U}(-\pi, \pi))
\end{equation*}
where $\mathcal{U}$ is the uniform distribution function, and random linear and angular speeds given by
\begin{equation*}
    (s,\; \dot{\theta}_{c}) = (\mathcal{U}(0, s_{lim}),\; \mathcal{U}(-\dot{\theta}_{lim}, \dot{\theta}_{lim})).
\end{equation*}
The parameter values used for the experimental setup are: $\alpha=0.5$, $x_\mathit{goal}=0$, $y_{goal}=10$, $\epsilon=0.05$, $x_{lim}=10$, $y_{lim}=10$, $s_{lim}= 2$ units/s, and $\dot{\theta}_{lim}=\frac{\pi}{4}$ rads/s. Each collected datapoint was a tuple 
\[
  (x_\mathit{diff}, y_\mathit{diff}, s, \theta, \dot{\theta_{c}}, \mathit{occ}), 
 \] 
 where $x_\mathit{diff}$ and $y_\mathit{diff}$ are the relative horizontal and vertical distances between the robot and the collider, and $\mathit{occ}$ is the label specifying whether the two agents are on collision course ($\mathit{occ}=2$) or not ($\mathit{occ}=1$). The datapoints were normalised such that $x_\mathit{diff},\theta,\dot{\theta}_{c}\in[-1,1]$ and $y_\mathit{diff}\in[0,1]$. Multiple simulations were performed to collect $6000$ collision datapoints and $6000$ no-collision datapoints, and the mean times to complete a journey between two successive waypoints with and without collision were recorded.

\medskip
\noindent
\textbf{DNN.} We used 80\% of the collected datapoints to train a two-class DNN classifier with the architecture proposed by Ehlers.~\cite{ehlers17} This architecture comprises a fully-connected linear layer with 40 nodes, followed
by a MaxPool layer with pool size 4 and stride size 1, a fully-connected ReLU layer with 19 nodes, and a final fully-connected ReLU layer with 2 nodes. The DNN was implemented and trained using TensorFlow in Python, with a cross-entropy loss function, the Adam optimization algorithm,~\cite{kingmaB14} and the following hyperparameters: 100~epochs, batch size 128, and initial learning rate 0.005 set to decay to 0.0001. 

\medskip
\noindent
\textbf{DeepDECS Stage 1: DNN uncertainty quantification.} We obtained the DNN uncertainty quantification probabilities~\eqref{eq:DNN-probabilities} using a test dataset comprising the 20\% of the dataset not used for training, and all possible combinations of the two DNN verification techniques described below.

\smallskip
\noindent
\emph{Minimum confidence threshold --} A $K$-class DNN classifier~\eqref{eq:classifier} maps its input $x \in \mathbb{R}^d$ to a discrete probability distribution $\delta(x)=(p_1,p_2,\ldots,p_K)$ over the $K$ classes, and outputs the class $\mathrm{argmax}_{i=1}^K\: p_i$  as the classifier prediction. As $\delta(x)$ is typically a poor estimate of the true probability distribution of $x$, we calibrated the DNN using the temperature scaling mechanism introduced by Guo et al.~\cite{GuoPSW17} as implemented by Kueppers et al.~\cite{Kueppers_2020_CVPR_Workshops}, and we defined the following verification method for the calibrated DNN: 
\begin{equation}
  \label{eq:confidence}
  \mathit{verif}_1(f,x) = 
  \left\{\begin{array}{ll}
  \hspace*{-1.5mm}\mathsf{true}, & \hspace*{-2mm}\textrm{if }
  \max_{i=1}^K\; p_i \geq \tau\\
  \hspace*{-1.5mm}\mathsf{false}, & \hspace*{-2mm}\textrm{otherwise}
  \end{array}\right.
\end{equation}
where $\tau$ is a threshold that we set to $0.8$.

\smallskip
\noindent
\emph{Local robustness certification} -- A DNN classifier~\eqref{eq:classifier} is $\epsilon$-locally robust at an input $x$ if perturbations within a small distance $\epsilon>0$ from $x$ (measured using the $\ell_2$ metric) do not lead to a change in the classifier prediction. For the second verification technique, we adopted the GloRo Net framework of Leino et al.~\cite{leino21gloro}. Given a DNN, this framework adds a network layer that augments the DNN with a local-robustness output by computing the Lipschitz constant of the function denoted by the original DNN and using it to verify local robustness. We added this GloRo Net layer to the trained collision prediction DNN, and we defined the following verification method for the augmented DNN:
\begin{equation}
  \label{eq:local}
  \mathit{verif}_2(f,x) = 
  \left\{\begin{array}{ll}
  \hspace*{-1.5mm}\mathsf{true}, & \hspace*{-2mm}\textrm{if }
  \forall x' \in \mathbb{R}^d.~||x-x'||_2 \leq \epsilon \\
  & \qquad\qquad\quad \implies f(x) = f(x')\\
  \hspace*{-1.5mm}\mathsf{false}, & \hspace*{-2mm}\textrm{otherwise}
  \end{array}\right.
\end{equation}
with the distance $\epsilon$ set to $0.05$.

\medskip
\noindent
\textbf{DeepDECS Stage 2: Model augmentation.} We developed a perfect-perception pDTMC model for the robot by instantiating the generic \acronym\ pDTMC from Figure~\ref{subfig:generic-DNN-pDTMC}. We implemented a software tool that automates the \acronym\ model augmentation process, and used this tool to obtain the DNN-perception pDTMC model for each combination of the verification methods~\eqref{eq:confidence} and~\eqref{eq:local} from the perfect-perception pDTMC and the DNN uncertainty quantification probabilities~\eqref{eq:DNN-probabilities} obtained in stage~1. The pDTMC models and the tool are presented in the supplementary material, and available on our project website.~\cite{DeepDECS-website}

\medskip
\noindent
\textbf{DeepDECS Stage 3: Controller synthesis.} As described earlier in the paper, probabilistic controllers were obtained by exploring the controller design space through discretisation. 

\medskip
\noindent
\textbf{Pareto front evaluation.} The quality of the controller Pareto fronts was evaluated (Figure~\ref{subfig:collision-limitation-Pareto-fronts-evaluation}) using the two metrics described below.

\smallskip
\noindent
\emph{Inverted Generational Distance~\cite{veldhuizen1999multiobjective} (IGD) -- } IGD measures the distance between the analysed Pareto front and a \emph{reference frame} (e.g., the true Pareto front, the best known approximation of the true Parero front, or an ``ideal'' Pareto front) by calculating, for each point on the reference frame, the distance to the closest point on the Pareto front. The IGD measure for the front is then computed as the mean of these distances. Smaller IGD values indicate better Pareto fronts. The IGD values from Figure~\ref{subfig:collision-limitation-Pareto-fronts-evaluation} were computed using the perfect-perception Pareto front as the reference frame.

\smallskip
\noindent
\emph{Hypervolume~\cite{zitzler1998multiobjective} (HV) -- } HV captures the proximity of the analysed Pareto front to a reference frame and the diversity of its points (where higher diversity is better) by measuring the volume (or area for two-dimensional Pareto fronts) delimited by these points and a reference point defined with respect to the reference framework. The HV values from Figure~\ref{subfig:collision-limitation-Pareto-fronts-evaluation} were obtain using the perfect-perception Pareto front as the reference frame and, in line with common practice,~\cite{10.1007/978-3-540-70928-2_64} its \emph{nadir} (i.e., the point corresponding to the worst values for each optimisation objective) as the reference point.

\bigskip
\noindent
\textbf{Driver-attentiveness management}

\medskip
\noindent
\textbf{Dataset.} The dataset for training the DNN and quantifying its uncertainty was taken from a user study~\cite{pakdamanian2021deeptake} conducted as part of our SafeSCAD project on the safety of shared control in autonomous driving.\cite{SafeSCAD} Each datapoint included: (i)~driver biometrics (eye movement, heart rate, and galvanic skin response); (ii)~driver gender; (iii)~driver perceived workload and psychological stress (estimated using established metrics); (iv)~driver engagement in non-driving tasks while not in control of the car (e.g., using a mobile phone, or reading); and (v)~vehicle data (distances to adjacent lanes and to any hazard that might be present, steering wheel angle, velocity, and gas and break pedal angles).

\medskip
\noindent
\textbf{DNN.} We used 60\% of the collected dataset for training a three-class DNN classifier with the architecture proposed by Pakdamanian et al.~\cite{pakdamanian2021deeptake}, and 15\% for its calibration and validation.

\medskip
\noindent
\textbf{DeepDECS Stage 1: DNN uncertainty quantification.} We obtained the DNN uncertainty quantification probabilities~\eqref{eq:DNN-probabilities} using a test dataset comprising the 25\% of the dataset not used for the DNN training, calibration and validation, and the two DNN verification techniques from~\eqref{eq:confidence} and~\eqref{eq:local}.

\medskip
\noindent
\textbf{DeepDECS Stage 2: Model augmentation} We devised a perfect-perception pDTMC model (provided as supplementary material) for the driver-attentiveness management system, and used our DeepDECS model augmentation tool to derive the DNN-perception pDTMC model for each setup (i.e., combination of verification methods from the DNN uncertainty quantification stage of \acronym).

\medskip
\noindent
\textbf{DeepDECS Stage 3: Controller synthesis.} The probabilistic model synthesis tool EvoChecker~\cite{gerasimou2018synthesis} was used to generate approximations of the Pareto-optimal set of probabilistic controllers. EvoChecker performs this synthesis using a multi-objective genetic algorithm (MOGA) whose fitness function is computed with the help of a probabilistic model checker. For all setups, we configured EvoChecker to use the NSGA-II MOGA with a population size of 1000 and a maximum number of evaluations set to $20\times10^{4}$, and the model checker PRISM. The EvoChecker executions were carried out using five CPUs and 8GB of memory on the University of York's Viking high-performance cluster (\url{https://www.york.ac.uk/it-services/services/viking-computing-cluster/#tab-1}), with a set maximum time of five hours. 

\medskip
\noindent
\textbf{Pareto front evaluation.} The quality of the approximate controller Pareto fronts was evaluated (Figure~\ref{subfig:SafeSCAD-Pareto-fronts-evaluation}) using the IGD and HV metrics described earlier.

\section*{Acknowledgements}

This project has received funding from the UKRI project EP/V026747/1 `Trustworthy Autonomous Systems Node in Resilience', the UKRI Global Research and Innovation Programme, and the Assuring Autonomy International Programme. The authors are grateful to the developers of the DeepTake deep neural network~\cite{pakdamanian2021deeptake} for sharing the DeepTake data sets, and to the University of York's Viking research computing cluster team for providing access to their systems.

\bibliographystyle{unsrt}
%\bibliography{ref.bib}

\end{small}

\newpage
~
\newpage
\appendix
\section{Supplementary information}

This supplementary information includes:
\begin{itemize}
    \item Presentations of the pDTMC models used in the mobile-robot collision limitation and driver-attentiveness management applications.
    \item A description of the tool that we implemented to automate the model augmentation stage of DeepDECS.
\end{itemize}

\subsection{pDTMC models for the mobile-robot\\ collision limitation application}

\begin{figure*}
    \centering
     \begin{subfigure}[c]{0.46\linewidth}
         \centering
         \includegraphics[width=1.0\linewidth]{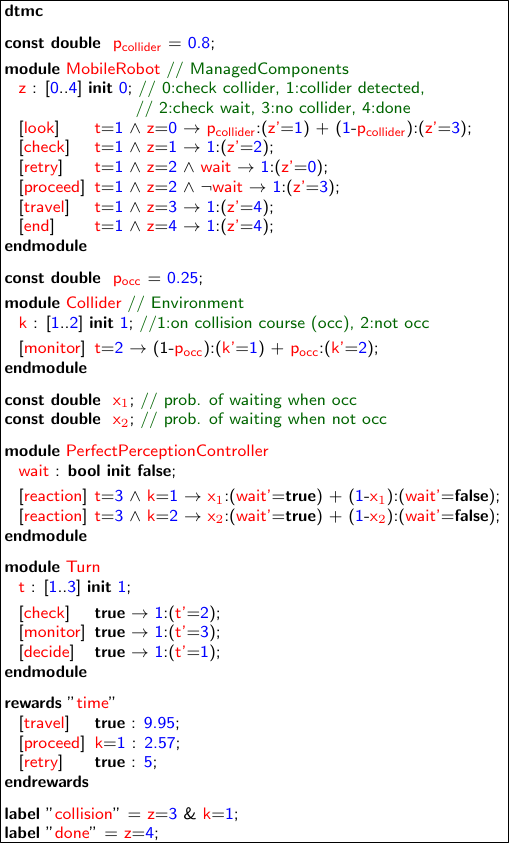}
         \caption{Perfect-perception pDTMC model of the mobile robot journey between two successive waypoints.  The reward structure models the \textcolor{red}{time} taken by each mobile robot actions: 9.95 time units are required (on average) to \textcolor{red}{travel} between adjacent waypoints (without collision), 2.57 additional time units are required when the robot decides to \textcolor{red}{proceed} despite being on a collision course, and five time units are required when the robot decides to \textcolor{red}{retry} after a short wait; the other robot actions are assumed to take negligible time. Two atomic propositions are used by the labelling function at the end of the model: \textcolor{red}{collision}, for states in which the robot travels despite being on collision course, and \textcolor{red}{done}, for states that mark the end of the journey to the next waypoint.}
         \label{subfig:collision-limitation-ideal-pDTMC}
     \end{subfigure}
     \hspace*{2mm}
     \begin{subfigure}[c]{0.502\linewidth}
         \centering
         \includegraphics[width=1.0\linewidth]{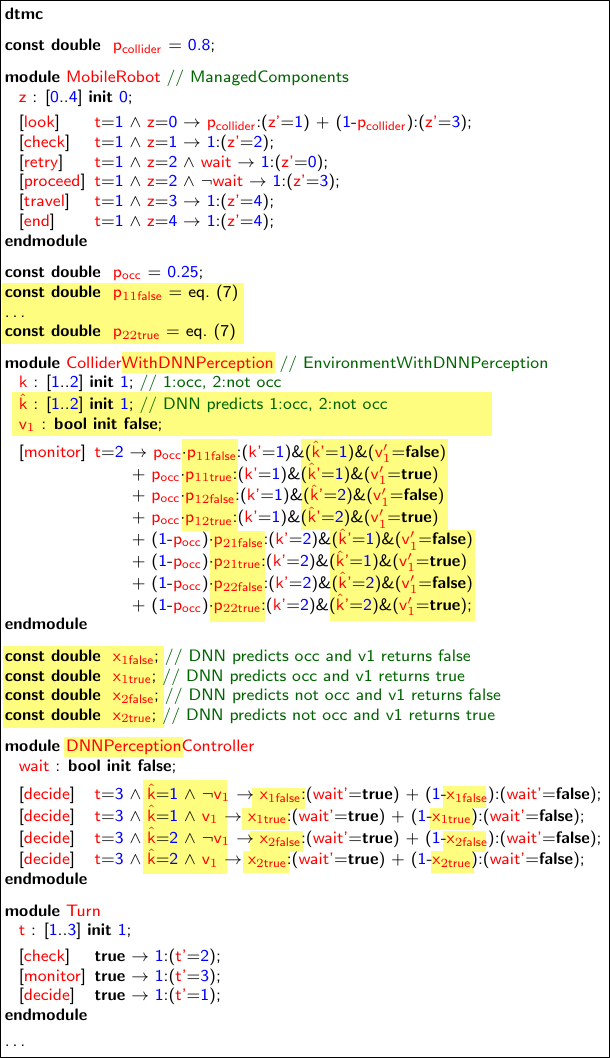}
         \caption{DNN-perception pDTMC model of the mobile robot journey between two successive waypoints. The probabilities $p_{kk'v_1}$ from the \textcolor{red}{ColliderWithDNNPerception} module quantify the DNN accuracy for ``verified'' inputs ($v_1=\mathsf{true}$) and ``unverified'' inputs ($v_1=\mathsf{false}$), and are used to model the class $\hat{k}$ predicted by the DNN when the true class is $k$. The decisions of the four-parameter probabilistic controller depend on the DNN prediction $\hat{k}$ and the online verification result $v_1$.}
         \label{subfig:collision-limitation-DNN-pDTMC}
     \end{subfigure}
     
     \vspace*{-2mm}
    \caption{\acronym\ pDTMC models from the robot collision limitation application}
    \label{fig:collision_pdtmc}
    
    \vspace*{-5mm}
\end{figure*}

\textbf{Perfect-perception pDTMC.} The logic underpinning the operation of the robot at any intermediate waypoint I from Figure~\ref{fig:collision-limitation} is modelled by the perfect-perception pDTMC in Figure~\ref{subfig:collision-limitation-ideal-pDTMC}. The model states are tuples 
\begin{equation}
  (z,k,t,\mathit{wait})\in \{0,1,\ldots,4\}\times [2]\times [3]\times \mathbb{B}
\end{equation} 
with the semantics from~\eqref{eq:state-partition}, where $k=1$ and $k=2$ correspond to the mobile robot not being on a collision course, and being on a collision course, respectively.

As shown by the \textcolor{red}{MobileRobot} pDTMC module, when reaching waypoint I the robot first uses its sensors (lidar, cameras, etc.) to \textcolor{red}{look} for the ``collider'' (state $z=0$). If the collider is present in the vicinity of the robot (which happens with probability $\mathsf{p_{collider}}$, known from previous executions of the task), the robot performs a \textcolor{red}{check} action (state $z=1$). As defined in the module \textcolor{red}{Collider}, this leads to the execution of a \textcolor{red}{monitor} action to predict whether travelling to the next waypoint would place the robot on collision course with the other agent (which happens with probability $\mathsf{p_{occ}}$, also known from historical data) or not. Each \textcolor{red}{monitor} action activates the controller, whose behaviour is specified by the \textcolor{red}{PerfectPerceptionController} module. A probabilistic controller with two parameters is used: the controller decides that the robot should wait with probability $x_1$ when no collision is predicted ($k=1$) and with probability $x_2$ if a collision is predicted ($k=2$). Depending on this decision, the robot will either \textcolor{red}{retry} after a short wait or \textcolor{red}{proceed} and \textcolor{red}{travel} to the next waypoint, reaching the \textcolor{red}{end} of the decision-making process. Finally, when the collider is absent (with probability $1-\mathsf{p_{collider}}$ in the first line from the \textcolor{red}{MobileRobot} module), the robot can \textcolor{red}{travel} without going through these intermediate actions.

\medskip
\noindent
\textbf{DNN-perception pDTMC.} Figure~\ref{subfig:collision-limitation-DNN-pDTMC} shows DNN-perception pDTMC model produced by the model augmentation \acronym\ stage when a single (generic) verification method is used in the DNN uncertainty quantification stage of \acronym. The DNN-perception pDTMC models for all uncertainty quantification options explored in our experiments (no verification method used, verification method $v_1$ from \eqref{eq:confidence} used, verification method $v_2$ from \eqref{eq:local} used, and both $v_1$ and $v_2$ used) are provided on our project website.

\begin{figure*}
     \centering
     \includegraphics[width=\linewidth]{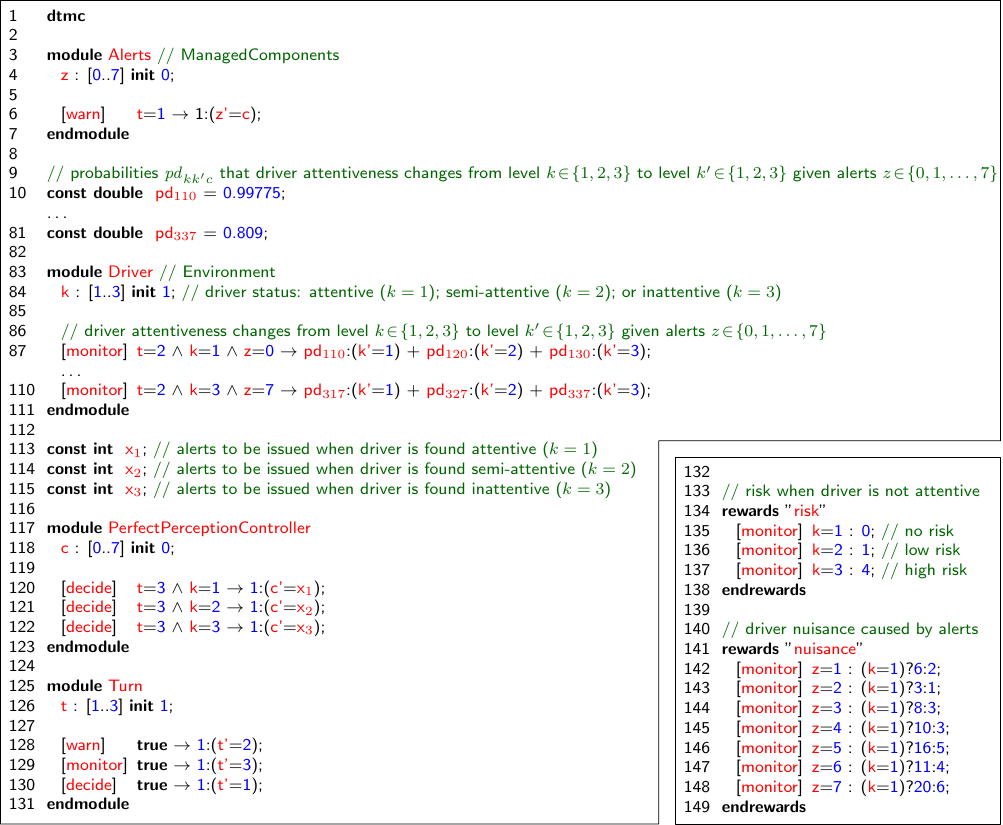}
     \caption{Perfect-perception pDTMC model of the SafeSCAD system. 
      The reward structures from lines 134--138 and 141--149 quantify the risk and driver nuisance associated with the different driver attentiveness levels and alert combinations, respectively. The expressions `$(k=1)?\mathit{value}_1:\mathit{value}_2$' from lines 142--148 evaluate to the larger $\mathit{value}_1$ if the driver is attentive (i.e., $k=1$), and $\mathit{value}_2$ otherwise.}
     \label{fig:SafeSCAD-model}
\end{figure*}

\subsection{pDTMC models for the driver\\ attentiveness management application}

\textbf{Perfect-perception pDTMC.} The operation of the driver attentiveness management system from Figure~\ref{fig:SafeSCAD} is modelled by the perfect-perception pDTMC in Figure~\ref{fig:SafeSCAD-model}. The pDTMC states are tuples 
\begin{equation}
    (z,k,t,c)\in[7]\times[3]^2\times\{0,1,\ldots,7\}
\end{equation}
with the semantics from~\eqref{eq:state-partition}, where the classes $k=1$, $k=2$ and $k=3$ correspond to the driver being attentive, semi-attentive and inattentive, respectively. 

The \textcolor{red}{Alerts} pDTMC module is responsible for \textcolor{red}{warn}ing the driver by  ``implementing'' the controller-decided alerts $c$. The \textcolor{red}{Driver} module models the driver attentiveness level $k$, which is \textcolor{red}{monitor}ed every 4s; the probabilities of transition between attentiveness levels depend on the combination of alerts $z$ in place. These probabilities are calculated using historical data. Each \textcolor{red}{monitor} action activates the controller, whose behavior is specified by the \textcolor{red}{PerfectPerceptionController} module. A deterministic controller is used; the control parameters $x_1,x_2,x_3\in\{0,1,\ldots,7\}$ are binary encodings of the alerts to be activated for each of the three driver attentiveness levels, e.g., $x_3=5=101_{(2)}$ corresponds to a deterministic-controller decision to have the optical alert active, the acoustic alert inactive, and the haptic alert active when the driver is inattentive. We note that the commands from the \textcolor{red}{PerfectPerceptionController} module are simplified representations of the following instantiations of the generic controller commands from the perfect-perception pDTMC template in Figure~\ref{subfig:generic-ideal-pDTMC}:\\[2mm]
\hspace*{3mm}[\textcolor{red}{decide}]\hspace*{2mm} \textcolor{red}{t}=\textcolor{blue}{3} $\wedge$ \textcolor{red}{k}=\textcolor{blue}{1} $\rightarrow$  
\textcolor{red}{x$_{10}$}:(\textcolor{red}{c'}=\textcolor{blue}{0}) + 
\textcolor{red}{x$_{11}$}:(\textcolor{red}{c'}=\textcolor{blue}{1}) + 
\textcolor{red}{x$_{12}$}:(\textcolor{red}{c'}=\textcolor{blue}{2}) +\\ \hspace*{25mm}
\textcolor{red}{x$_{13}$}:(\textcolor{red}{c'}=\textcolor{blue}{3}) + 
\textcolor{red}{x$_{14}$}:(\textcolor{red}{c'}=\textcolor{blue}{4}) + 
\textcolor{red}{x$_{15}$}:(\textcolor{red}{c'}=\textcolor{blue}{5}) + \\ \hspace*{25mm}
\textcolor{red}{x$_{16}$}:(\textcolor{red}{c'}=\textcolor{blue}{6}) + 
\textcolor{red}{x$_{17}$}:(\textcolor{red}{c'}=\textcolor{blue}{7});\\
\hspace*{3mm}[\textcolor{red}{decide}]\hspace*{2mm} \textcolor{red}{t}=\textcolor{blue}{3} $\wedge$ \textcolor{red}{k}=\textcolor{blue}{2} $\rightarrow$  
\textcolor{red}{x$_{20}$}:(\textcolor{red}{c'}=\textcolor{blue}{0}) + 
\textcolor{red}{x$_{21}$}:(\textcolor{red}{c'}=\textcolor{blue}{1}) + 
\textcolor{red}{x$_{22}$}:(\textcolor{red}{c'}=\textcolor{blue}{2}) + \\ \hspace*{25mm}
\textcolor{red}{x$_{23}$}:(\textcolor{red}{c'}=\textcolor{blue}{3}) + 
\textcolor{red}{x$_{24}$}:(\textcolor{red}{c'}=\textcolor{blue}{4}) + 
\textcolor{red}{x$_{25}$}:(\textcolor{red}{c'}=\textcolor{blue}{5}) + \\ \hspace*{25mm}
\textcolor{red}{x$_{26}$}:(\textcolor{red}{c'}=\textcolor{blue}{6}) + 
\textcolor{red}{x$_{27}$}:(\textcolor{red}{c'}=\textcolor{blue}{7});\\
\hspace*{3mm}[\textcolor{red}{decide}]\hspace*{2mm} \textcolor{red}{t}=\textcolor{blue}{3} $\wedge$ \textcolor{red}{k}=\textcolor{blue}{3} $\rightarrow$  
\textcolor{red}{x$_{30}$}:(\textcolor{red}{c'}=\textcolor{blue}{0}) + 
\textcolor{red}{x$_{31}$}:(\textcolor{red}{c'}=\textcolor{blue}{1}) + 
\textcolor{red}{x$_{32}$}:(\textcolor{red}{c'}=\textcolor{blue}{2}) + \\ \hspace*{25mm}
\textcolor{red}{x$_{33}$}:(\textcolor{red}{c'}=\textcolor{blue}{3}) + 
\textcolor{red}{x$_{34}$}:(\textcolor{red}{c'}=\textcolor{blue}{4}) + 
\textcolor{red}{x$_{35}$}:(\textcolor{red}{c'}=\textcolor{blue}{5}) + \\ \hspace*{25mm}
\textcolor{red}{x$_{36}$}:(\textcolor{red}{c'}=\textcolor{blue}{6}) + 
\textcolor{red}{x$_{37}$}:(\textcolor{red}{c'}=\textcolor{blue}{7});\\[2mm]
This simplification is possible because we are synthesising deterministic controllers and therefore, in addition to $\sum_{i=0}^7 x_{ki}=1$ for every $k\in \{1,2,3\}$, we have $\forall i\in \{0,1,\ldots,7\}: x_{ki}\in\{0,1\}$.

\begin{figure*}
     \centering
     \includegraphics[width=\linewidth]{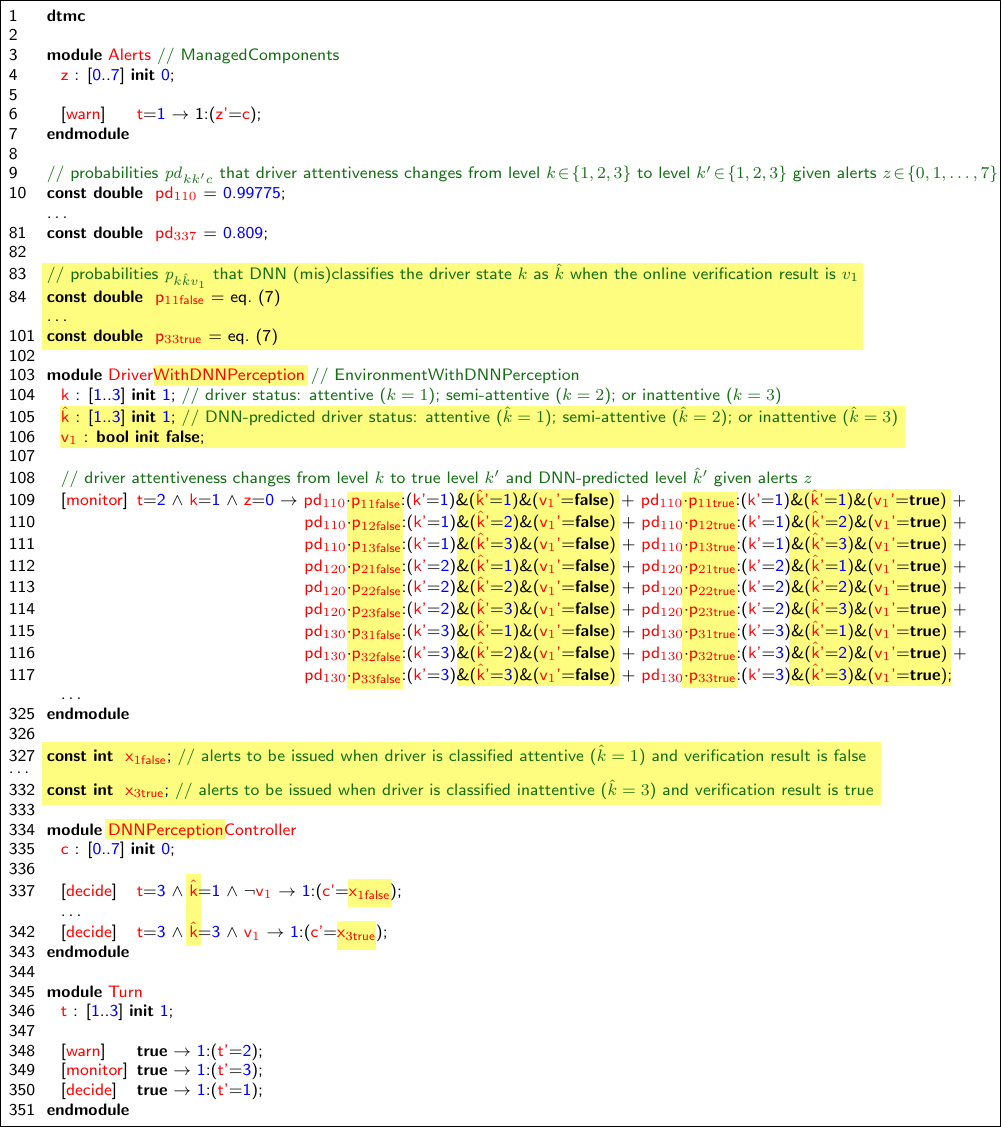}
     \caption{DNN-perception pDTMC model of the SafeSCAD driver-attentiveness management system for the scenario when a single DNN verification method is used to distinguish between ``verified'' ($v_1=\mathsf{true}$) and ``unverified'' ($v_1=\mathsf{false}$) DNN predictions of the driver's attentiveness level. Lines~109--117 show how all combinations of true ($k'$) and DNN-predicted ($\hat{k}'$) driver attentiveness levels that can be reached from the attentive driver state ($k=1$) when no alerts are used ($z=0$). The six-parameter deterministic controller decides a combination of alerts $c$ for each pair of DNN-predicted driver atentiveness level $\hat{k}$ and online DNN verification result $v_1$ (lines 337--342). The \textcolor{red}{Turn} module is unchanged, and the two reward structures from the pDTMC in Figure~\ref{fig:SafeSCAD-model} are omitted for brevity. 
     \label{fig:SafeSCAD-DNN-model}}
\end{figure*}

\medskip
\noindent
\textbf{DNN-perception pDTMC.} Figure~\ref{fig:SafeSCAD-DNN-model} shows DNN-perception pDTMC model produced by the model augmentation \acronym\ stage when a single (generic) verification method is used in the DNN uncertainty quantification stage of \acronym. The DNN-perception pDTMC models for all uncertainty quantification options explored in our experiments (no verification method used, verification method $v_1$ from \eqref{eq:confidence} used, verification method $v_2$ from \eqref{eq:local} used, and both $v_1$ and $v_2$ used) are provided on our project website.

\vspace*{-3mm}
\subsection{DeepDECS model augmentation tool \label{sect:tool}}
A python tool was developed to automate the model augmentation stage of DeepDECS. The tool takes as input a perfect-perception pDTMC model with the structure from Figure~\ref{subfig:generic-ideal-pDTMC} and the confusion matrices~\eqref{eq:confusion} and outputs a DNN-perception pDTMC model with the structure from Figure~\ref{subfig:generic-DNN-pDTMC}.
The tool is executed as\\[2mm]
\hspace*{2mm}\texttt{python deepDECSAugment.py perfect-perc-model.pm\\
\hspace*{15mm} confusion\_matrices.txt DNN-perc-model.pm}

\noindent
where:
\begin{itemize}
    \item \texttt{perfect-perc-model.pm} is a file containing the perfect-perception pDTMC model;
    \item \texttt{confusion\_matrices.txt} is a file containing  the confusion matrix elements $\mathcal{C}_v[k,k']$, $v\in \mathbb{B}^n$, $k\in[K]$, $k'\in[K]$, from~\eqref{eq:confusion} in the format:
\[\begin{array}{l l l}
    \mathcal{C}_{(\mathsf{false},\mathsf{false},\ldots,\mathsf{false})}[1,1] &  \cdots & \mathcal{C}_{(\mathsf{false},\mathsf{false},\ldots,\mathsf{false})}[1,K] \\
    \mathcal{C}_{(\mathsf{false},\mathsf{false},\ldots,\mathsf{false})}[2,1] &  \cdots & \mathcal{C}_{(\mathsf{false},\mathsf{false},\ldots,\mathsf{false})}[2,K] \\
    \cdots\\
    \mathcal{C}_{(\mathsf{false},\mathsf{false},\ldots,\mathsf{false})}[K,1] & \cdots & \mathcal{C}_{(\mathsf{false},\mathsf{false},\ldots,\mathsf{false})}[K,K] \\
\end{array}
\]

\vspace*{-6mm}
\[\begin{array}{l l l}
    \mathcal{C}_{(\mathsf{true},\mathsf{false},\ldots,\mathsf{false})}[1,1] &  \cdots & \mathcal{C}_{(\mathsf{true},\mathsf{false},\ldots,\mathsf{false})}[1,K] \\
    \mathcal{C}_{(\mathsf{true},\mathsf{false},\ldots,\mathsf{false})}[2,1] &\cdots & \mathcal{C}_{(\mathsf{true},\mathsf{false},\ldots,\mathsf{false})}[2,K] \\
    \cdots\\
    \mathcal{C}_{(\mathsf{true},\mathsf{false},\ldots,\mathsf{false})}[K,1] &  \cdots & \mathcal{C}_{(\mathsf{true},\mathsf{false},\ldots,\mathsf{false})}[K,K] \\
    \cdots\\
    \mathcal{C}_{(\mathsf{true},\mathsf{true},\ldots,\mathsf{true})}[1,1] &  \cdots & \mathcal{C}_{(\mathsf{true},\mathsf{true},\ldots,\mathsf{true})}[1,K] \\
    \mathcal{C}_{(\mathsf{true},\mathsf{true},\ldots,\mathsf{true})}[2,1] &  \cdots & \mathcal{C}_{(\mathsf{true},\mathsf{true},\ldots,\mathsf{true})}[2,K] \\
    \cdots\\
    \mathcal{C}_{(\mathsf{true},\mathsf{true},\ldots,\mathsf{true})}[K,1] &  \cdots & \mathcal{C}_{(\mathsf{true},\mathsf{true},\ldots,\mathsf{true})}[K,K] \\
\end{array}
\]
  \item \texttt{DNN-perc-model.pm} is the name of the file in which the DNN-perception pDTMC model will be generated.
\end{itemize}

\end{document}